\documentclass[manuscript,screen]{acmart}
\AtBeginDocument{%
  }

\usepackage{booktabs}
\usepackage{tabularx}
\usepackage{xltabular} 
\usepackage{array}
\usepackage{multirow}
\usepackage{threeparttable}
\usepackage{subfig}

\DeclareGraphicsExtensions{.pdf,.jpeg,.jpg,.png}

\setcopyright{acmlicensed}
\acmJournal{CSUR}
\acmYear{2026} \acmVolume{1} \acmNumber{1} \acmArticle{1} \acmMonth{1} 
\acmDOI{xxx}

\acmJournal{JACM}
\acmVolume{0}
\acmNumber{0}
\acmMonth{8}

\begin{document}

\title{Federated Prompt Learning: A Unified Framework, Empirical Analysis, and Future Directions}

\author{Qinglin Yang}
\orcid{0000-0002-7263-8914}
\email{yangqinglin@gzhu.edu.cn}
\affiliation{%
  \institution{Cyberspace Institute of Advanced Technology, Guangzhou University; Guangdong Key Laboratory of Industrial Control System Security; Huangpu Research School of Guangzhou University}
  \city{Guangzhou}
  \state{Guangdong}
  \country{China}
}

\author{Chen Qiu}
\email{c-qiu@ibrc.or.jp}
\affiliation{%
  \department{Department of Horticultural Science}
  \institution{Iwate Biotechnology Research Center}
  \city{Iwate}
  \country{Japan}
}

\author{Hongyuan Zhang}
\email{b23110420@njupt.edu.cn}
\affiliation{%
  \department{Computer Science and Technology}
  \institution{Nanjing University of Posts and Telecommunications}
  \city{Nanjing}
  \state{Jiangsu}
  \country{China}
}

\author{Pengdeng Li}, \author{Yuan Liu} \authornote{The Corresponding author.}, \author{Zhihong Tian}
\email{\{pdli, yuanliu,tianzhihong\}@gzhu.edu.cn}
\affiliation{%
  \institution{Cyberspace Institute of Advanced Technology, Guangzhou University; Guangdong Key Laboratory of Industrial Control System Security; Huangpu Research School of Guangzhou University}
  \city{Guangzhou}
  \state{Guangdong}
  \country{China}
}

\renewcommand{\shortauthors}{Q.Yang C.Qiu, H.Zhang et al.}

\begin{abstract}
Large language models (LLMs) have become core components of cloud-based intelligent services in academia and industry, yet their training and deployment are hindered by high computational costs, data centralization, and privacy concerns. Federated learning (FL) offers a decentralized training paradigm that enables clients to collaboratively train a learning model without sharing raw data, making it a promising solution for privacy-preserving LLM training and reasoning. This paper presents a comprehensive survey of federated prompt learning (FPL) to review recent advances in integrating the federated learning paradigm and large language models, answering the following research questions: RQ1: The fundamental motivations, characteristics, and enabling technologies of FPL, and how it differs from conventional FL and full-model federated fine-tuning; RQ2: The trade-offs FPL approaches exhibit in performance, communication efficiency, computational overhead, scalability, personalization, and heterogeneity handling; RQ3: The remaining security, privacy, robustness, and system challenges, along with key future research directions. To this end, we systematically examine existing FPL methods across the full model lifecycle: pre-training, fine-tuning, and practical applications, while discussing security, privacy, and robustness issues and summarizing existing defense mechanisms. Finally, we highlight open challenges and future directions, aiming to help readers understand how the insights drive research in FPL.
\end{abstract}



\maketitle

\section{Introduction}
Large language models (LLMs) have become the cornerstone of modern artificial intelligence, demonstrating remarkable capabilities in natural language understanding, reasoning, and generation across a wide range of domains, including healthcare, finance, software engineering, and education~\cite{chang2024survey}. 
However, the rapid scaling of LLMs has exposed fundamental limitations of the prevailing centralized training and deployment paradigm. For training, the reliance on massive centralized datasets raises severe privacy and regulatory concerns, while the growing scarcity of high-quality public data and the prohibitive computational and communication costs increasingly constrain further model scaling.
These challenges motivate the exploration of new collaborative and privacy-preserving learning paradigms for the development and adaptation of LLMs.

FL has emerged as a promising decentralized framework that enables multiple data owners to collaboratively train machine learning models without sharing raw data~\cite{zhang2021survey}. When FL meets LLMs, it raises some diverse challenges compared with conventional FL patterns.
(i) The distinctive significance of FL for LLMs is the stronger demand for multi-source heterogeneous data.
(ii) Conventional LLMs usually rely on centralized training, in which massive amounts of data are aggregated onto a unified platform before pre-training or fine-tuning is conducted. However, in real-world scenarios, many forms of high-value data cannot be centralized, such as internal enterprise documents, medical data, financial data, legal data, personal device data, and cross-institutional business data. 
(iii) In terms of model architecture, FL promotes the evolution of LLMs from a single unified model toward a structure in which shared capabilities and personalized capabilities coexist.

Therefore, the integration of LLMs with FL faces greater technical challenges than traditional deep learning. LLMs usually have enormous parameter scales, often reaching billions or even hundreds of billions of parameters, which is several orders of magnitude larger than conventional deep learning models (e.g., GoogLeNet, AlexNet, VGG, and ResNet).
If the full model parameters are directly transmitted, both communication costs and storage pressure become extremely high. Hence, the core issue is no longer merely how to aggregate models from multiple clients, but how to transmit and update only the most critical, minimal, and effective parameters under extremely large model scales.

To this end, prompt learning and parameter-efficient fine-tuning (PEFT) have gained significant attention as lightweight alternatives to full-model fine-tuning. By freezing the backbone of a pretrained foundation model and optimizing only a small number of task-specific parameters—such as soft prompts, prefixes, or low-rank adapters (LoRA)~\cite{hu2022lora}, training cost is drastically reduced while preserving the generalization power of large models.
Importantly, the compactness and modularity of prompt-based updates make them inherently well-suited to federated environments. This observation has led to a new paradigm, federated prompt learning (FPL), which leverages prompt learning as the primary interface between LLMs and federated optimization.

FPL fundamentally reshapes how LLMs are trained, adapted, and deployed in distributed settings. Instead of federating the entire model or large subsets of parameters, clients collaboratively optimize lightweight prompt or adapter modules on top of a shared frozen foundation model. 
This design dramatically reduces communication costs, mitigates client-side resource constraints, and enhances robustness to non-IID data, while maintaining strong privacy guarantees. In consequence, federated prompt learning has rapidly evolved from early proof-of-concept studies to a rich ecosystem of methods spanning prompt-based fine-tuning, personalization, split and off-site architectures, multimodal learning, and real-world applications.

\begin{figure*}[t]
    \centering
    \includegraphics[width=0.85\linewidth]{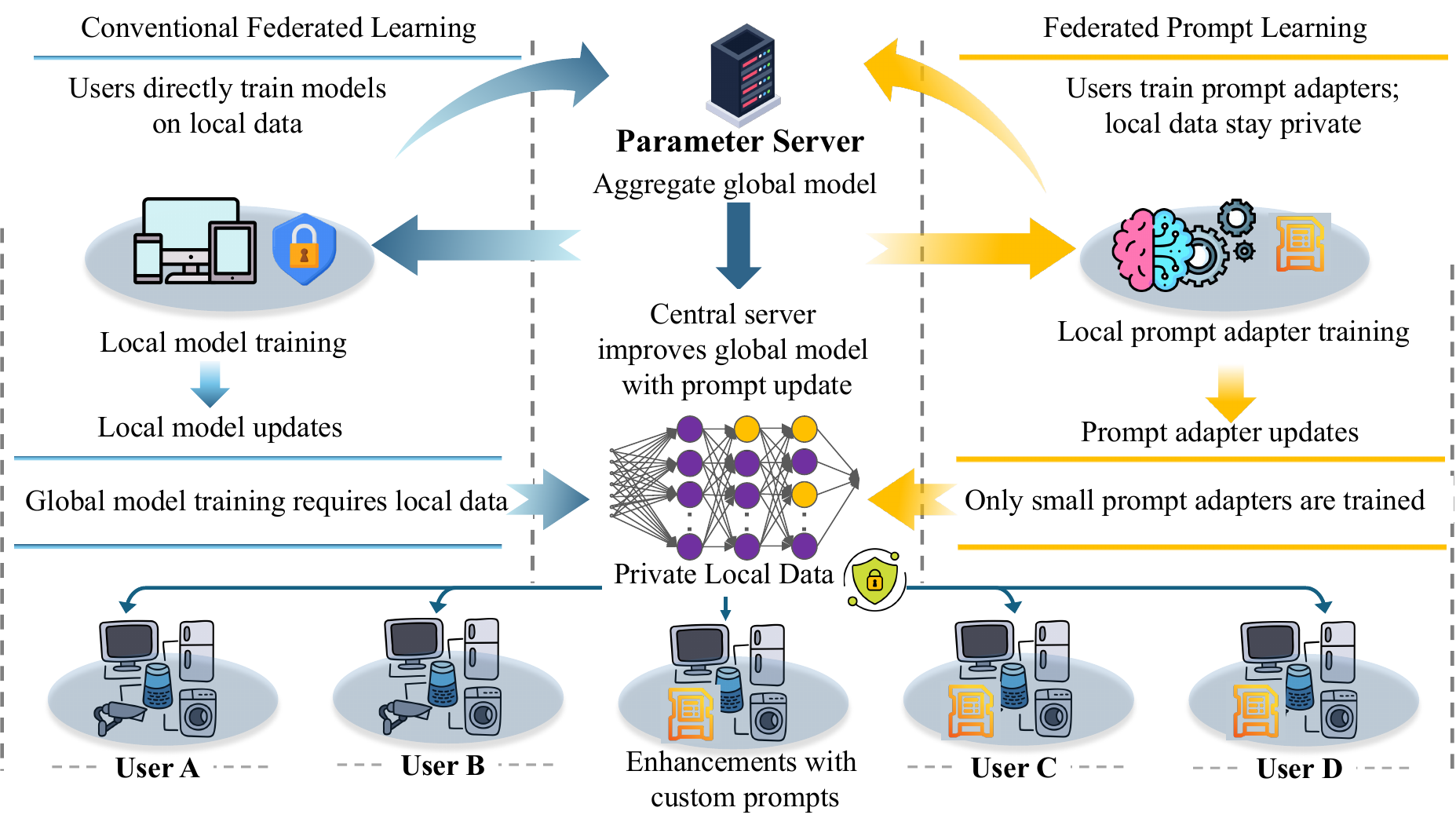}
    \caption{The Difference between conventional FL and federated prompt learning.}
    \Description{Side-by-side workflow comparison. Conventional federated learning trains local models and sends model updates to a parameter server. Federated prompt learning keeps the foundation model fixed, trains only small prompt adapters on private client data, and aggregates prompt updates to improve the global model.}
    \label{fig:FLComparison}
\end{figure*}

\subsection{Motivations and Contributions}
Despite this rapid progress, the existing literature remains fragmented. Prior surveys~\cite{villalobos2024position,fan2023fate,thakur2025analyzing,yao2024survey,yang2025synergizing,yan2025federated} have primarily focused on either federated learning in general, federated fine-tuning of LLMs, or the security and privacy of LLMs. Despite the rapid growth of this field, its literature remains fragmented across machine learning, distributed systems, cloud–edge computing, multimodal learning, and cybersecurity. Existing surveys typically examine general FedLLM architectures, federated fine-tuning, IoT-oriented deployments, or LLM security independently. Prompt- and adapter-based methods are often treated as auxiliary implementation techniques rather than as a distinct distributed optimization and service-delivery paradigm. Moreover, previous reviews rarely connect the complete model-service lifecycle, including training, personalization, inference, deployment, application, and protection, or systematically compare methods in terms of computational overhead, communication cost, scalability, dependability, and privacy.

A systematic and unified treatment that centers on federated prompt learning as a distinct paradigm, examines its role across the full LLM lifecycle, and synthesizes insights from algorithmic, system, application, and security perspectives is still lacking. Without such a comprehensive view, it is difficult for researchers and practitioners to understand the design trade-offs, identify open challenges, and navigate future research directions in this emerging field. To address this gap, this paper presents a comprehensive survey of FPL for LLMs that develops a unified, lifecycle-oriented understanding of federated prompt learning from both algorithmic and cloud-system perspectives. 
It organizes existing approaches by their technical mechanisms and deployment roles, and examines how prompt-based adaptation interacts with cloud coordination, heterogeneous edge resources, distributed inference, multimodal applications, and adversarial environments. Particular attention is given to performance, communication efficiency, client-side computation and memory, scalability, personalization, privacy, robustness, and practical deployability. It is critically important to answer the following research questions(RQs) comprehensively:
\begin{enumerate}
    \item \textbf{RQ1}: What are the fundamental motivations, characteristics, and enabling technologies of federated prompt learning, and how does it differ from conventional federated learning and full-model federated fine-tuning?
    \item \textbf{RQ2}: What trade-offs do federated prompt learning approaches exhibit in performance, communication efficiency, computational overhead, scalability, personalization, and heterogeneity handling?
    \item \textbf{RQ3}: What security, privacy, robustness, and system challenges remain, and what future research directions are most important?
\end{enumerate}
Our contributions are summarized as follows:
\begin{itemize}
    \item The core contribution of this survey lies in the experimental validation of representative federated prompt learning models and frameworks, through which we derive key empirical findings and practical insights. These results are intended to provide guidance and inspiration for researchers to advance research in this emerging area.
    \item We analyze the security, privacy, and robustness challenges that arise when prompt learning meets federated optimization, and summarize existing attack models and defense mechanisms tailored to federated LLMs.
    \item We distill key lessons from current research and outline open challenges and promising future directions to guide subsequent studies in federated prompt learning.
\end{itemize}
Through this survey, we aim to provide a structured and coherent reference for researchers and practitioners seeking to understand how prompt learning can effectively bridge large language models and federated learning, enabling scalable, efficient, and privacy-preserving collaboration in the era of foundation models.
\subsection{Methods }
This review combines a systematic literature review with taxonomy-based comparative analysis. Relevant studies are retrieved from databases, such as ``Web of Science", ``Google Scholar", and ``dblp", using predefined search terms related to federated learning and prompt learning. Explicit inclusion and exclusion criteria are applied according to publication period, document type, research relevance, and methodological completeness. After duplicate removal, title and abstract screening, and full-text assessment, 52 high-quality studies were retained for subsequent analysis.
The selected studies are then systematically coded in terms of research objectives, core methods, model architectures, datasets, evaluation metrics, empirical results, and reported limitations. Based on their underlying technical mechanisms, the reviewed approaches are organized into a structured taxonomy and compared with respect to performance, efficiency, communication cost, computational overhead, privacy protection, scalability, and application scenarios. Finally, the review synthesizes the major advances, unresolved challenges, and potential directions for future research.

\section{Related Work and Background}\label{sec:relatedwork}

\subsection{Related work}

Recent research has increasingly investigated the integration of LLMs and FL as a response to the escalating challenges of data scarcity, privacy preservation, communication cost, and the prohibitive cost of centralized model training. 
Nikolaou \emph{ et al.}~\cite{nikolaou2025language} establish a fundamental theoretical property of decoder-only Transformer language models, showing that they are almost surely injective and hence exactly invertible with respect to their input prompts.
Contrary to the prevailing intuition that nonlinearities, normalization, and attention mechanisms inevitably lead to information loss, the authors prove that distinct input prompts map to distinct hidden representations with probability one under standard architectures, continuous parameter initialization, and finite-step gradient-based training. 
Leveraging tools from real analysis, they show that collisions can only occur on measure-zero parameter sets and that common training procedures cannot reach such pathological configurations. 
Beyond theory, Nikolaou \emph{ et al.} introduce SIPIT, the first provably correct and efficient algorithm that reconstructs the exact input prompt from internal hidden states in linear time, exploiting the causal structure of Transformers. 

Cheng et al.~\cite{cheng2024towards} provide a comprehensive, structured overview of federated large language models (FedLLMs), elucidating motivations, methodologies, and future directions at the intersection of FL and LLMs. They highlight FedLLMs as a promising paradigm for addressing fundamental LLM challenges, including exhaustion of high-quality public data, stringent privacy requirements, continuous model updating, and prohibitive computational and communication costs.
By systematically reviewing the full FedLLM lifecycle, from pre-training and fine-tuning to deployment and application, the paper categorizes existing approaches such as parameter-efficient fine-tuning, prompt-based and split learning methods, personalized federated LLMs, and backpropagation-free techniques, analyzing their performance trade-offs under data, system, and model heterogeneity. The survey further identifies critical challenges, including communication bottlenecks, synchronization and straggler issues, non-IID data, and amplified security and privacy threats such as poisoning, backdoors, and inference attacks, and summarizes current defense mechanisms and their limitations.

For data scarcity, Villalobos et al. \cite{villalobos2024position} quantitatively demonstrate that the supply of high-quality public human-generated text is finite and likely to be exhausted by frontier LLMs between 2026 and 2032 under current scaling trends. This finding underscores a fundamental limitation of centralized data-driven scaling and highlights the necessity of leveraging decentralized, privately held data sources, thereby motivating privacy-preserving collaborative learning paradigms (e.g., FL).

On the systems side, Fan \emph{et al.} \cite{fan2023fate} propose FATE-LLM, an industrial-grade FedLLM framework that shows how parameter-efficient fine-tuning (PEFT), off-site tuning, and knowledge distillation can substantially reduce communication overhead while approaching centralized fine-tuning performance. These core techniques of the work demonstrate the practical feasibility of federated LLM training in enterprise environments, but primarily focus on system implementation rather than a unified methodological or conceptual abstraction. Chen~\emph{et al.}~\cite{chen2024integration} systematically review the emerging integration of LLMs and federated learning, highlighting their complementarity in addressing data scarcity, privacy constraints, and scalability. It organizes prior work into a unified framework spanning: (i) incorporating LLM sub-technologies, pre-training and prompt engineering, into FL to improve convergence speed, robustness to non-IID data, personalization, and domain generalization; (ii) applying FL sub-technologies, distributed computing and privacy-preserving mechanisms, to mitigate LLMs' high computational cost, limited data coverage, and privacy risks; and (iii) holistic federated LLM (FedLLM) systems that jointly optimize training, instruction tuning, and alignment under decentralized settings.

Several surveys have attempted to organize the rapidly growing literature at the intersection of LLMs and FL. For example, Thakur et al. \cite{thakur2025analyzing} provide a broad unifying analysis of FL–LLM fusion, categorizing prior work into paradigms that either use LLM capabilities to enhance FL, apply FL to improve LLM adaptation, or jointly design holistic FedLLM systems. Nevertheless, this review adopts a macro-level perspective and does not systematically analyze prompt- or adapter-based learning as a first-class federated optimization interface. Yao et al. \cite{yao2024survey} focus on security and privacy issues of LLMs, offering an extensive taxonomy of risks and defenses, but largely decouple these concerns from federated training dynamics. Yang et al. \cite{yang2025synergizing} further expand the scope by reviewing the joint integration of IoT, LLMs, and FL, emphasizing architectural synergies in edge systems, yet their analysis is application-driven and not centered on LLM adaptation mechanisms. At a finer granularity, Yan et al. \cite{yan2025federated} conduct a comparative study of federated fine-tuning paradigms for LLMs, including FedLLMs, KD-FedLLMs, and Split-FedLLMs. Their results reveal inherent trade-offs between model accuracy, communication cost, and client-side computation, illustrating that no single federated framework is universally optimal.

Despite these substantial advances, existing studies and surveys exhibit a notable gap: the lack of a dedicated, systematic treatment of FPL as a distinct and unifying paradigm for LLM–FL integration. Prompt- and adapter-based methods have emerged as the dominant practical mechanism for adapting large language models under resource, communication, and privacy constraints, yet they are often discussed only as auxiliary techniques within broader FedLLM frameworks or system implementations.

Different from the existing surveys, our review places FPL at the center of analysis, systematically reviewing how prompt- and adapter-based techniques enable efficient, scalable, and privacy-preserving collaboration of large language models under federated settings. By organizing existing methods along the full LLM lifecycle and jointly examining algorithmic, system-level, application-driven, and security-related dimensions, this survey constructs a unified framework that provides answers to the questions: \textit{Why do LLMs need FL?}, \textit{How does FPL work?}, \textit{What can FL and LLMs do for each other?}, \textit{What kinds of challenges can FPL meet?}, and \textit{What are the potential research directions of FPL?}

\subsection{Background and Fundamentals}

This subsection introduces the fundamental concepts underlying federated prompt learning. We first describe large language models and their adaptation requirements, followed by conventional federated learning, prompt learning, and parameter-efficient fine-tuning. We then formalize federated prompt learning as a lightweight interface for adapting foundation models using decentralized private data.

\subsubsection{Large Language Models and Model Adaptation}

LLMs are typically built on transformer~\cite{han2021transformer} architectures and pretrained on large-scale text corpora using self-supervised objectives. Through large-scale pretraining, LLMs acquire general-purpose representations that support a broad range of downstream capabilities, including language understanding, text generation, reasoning, question answering, and instruction following~\cite{chang2024survey}. These pretrained models can subsequently be adapted to specific tasks, domains, organizations, or users.

LLMs containing billions of parameters impose substantial computation, memory, communication, and storage requirements. In distributed settings, repeatedly transmitting full-model parameters or gradients further creates a major communication bottleneck. Full fine-tuning may also require each participant to possess sufficient hardware resources to store and optimize the complete model.
The adaptation of LLMs is further complicated by the decentralized nature of high-value data. Enterprise documents, medical records, financial information, source code, personal-device data, and institution-specific knowledge are frequently distributed across independent data owners. Privacy requirements, regulatory restrictions, and commercial confidentiality often prevent these data from being transferred to a centralized training platform. 
These requirements motivate the combination of federated learning with lightweight model-adaptation techniques.

\subsubsection{Conventional Federated Learning}
As illustrated in Fig.~\ref{fig:FLComparison}, a typical FL system consists of a central server and a set of participating clients. Each client maintains a local dataset and performs model optimization within its own trusted environment. The server coordinates training by distributing model parameters, collecting local updates, and aggregating them into a global model. A conventional FL round generally contains the following three steps: global model broadcast, local model training and transmission, and model aggregation~\cite{li2020federated}.

Let $K$ denote the number of clients, $D_{k}$
 the private dataset of client $k$, and $n_{k}=|D_{k}|$ its number of samples. Conventional FL can be expressed as the minimization of a weighted global objective:

\begin{equation}
\min _{\mathbf{w}} F(\mathbf{w})=\sum_{k=1}^K \frac{n_k}{\sum_{j=1}^K n_j} F_k(\mathbf{w}),
\end{equation}
where $w$ represents the shared model parameters and $F_{k}(\mathbf{w})$ is the local objective evaluated on $D_{k}$. In FedAvg, the server aggregates the locally optimized parameters as:
\begin{equation}
\mathbf{w}^{t+1}=\sum_{k \in S_t} \frac{n_k}{\sum_{j \in S_t} n_j} \mathbf{w}_k^{t+1},
\end{equation}
where $S_t$ is the set of clients participating in communication round $t$.

The clients in conventional FL are usually heterogeneous, including non-independent and identically distributed data, computation capacities, storage, and communication.
This case becomes more severe when FL is directly applied to LLMs. Transmitting and optimizing an entire LLM may exceed the computation, memory, and bandwidth available to many clients. Therefore, federated LLM systems generally require parameter-efficient adaptation, compression, model partitioning, or off-site training rather than straightforward full-model federated optimization.

\subsubsection{Prompt Learning} \label{sub2:Promptlearning}

The prompt plays a role in bridging the downstream task and the knowledge encoded in the pretrained model.
Prompt learning adapts a pretrained model by modifying or learning the conditioning context while keeping most or all backbone parameters frozen~\cite{zhou2022conditional}. This paper treats ``prompt tuning" as a class of methods rather than a particular method.

Prompts can generally be divided into discrete and continuous forms. A discrete prompt consists of human-readable tokens, task descriptions, demonstrations, or instructions inserted into the model input. Continuous prompt learning instead represents the prompt as a set of trainable embedding vectors. 

In prompt tuning, learnable embeddings are appended or prepended to the input sequence. Prefix tuning introduces trainable vectors into multiple transformer layers, commonly through the attention~\cite{vaswani2017attention} mechanism.  In vision or vision-language models, visual prompts may be introduced into image patches or intermediate visual representations, while textual prompts condition the language branch.

Let $f_{\theta}$ denote a pretrained model with frozen parameters $\theta$, and let $p$ denote a trainable prompt. Prompt learning optimizes:
\begin{equation}
\min _{\mathbf{p}} \mathcal{L}\left(f_{\boldsymbol{\theta}}(\mathbf{x} ; \mathbf{p}), y\right),
\end{equation}
where $(x,y)$ is a downstream training example. Because $\theta$ remains fixed and the dimension of $p$ is substantially smaller than that of the backbone, prompt learning reduces the computation, storage, and communication required for adaptation.

However, prompt learning remains sensitive to the quality of the pretrained model and the distance between the pretraining and downstream distributions.

\subsubsection{Parameter-Efficient Fine-Tuning}

Parameter-efficient fine-tuning (PEFT)~\cite{fu2023effectiveness} refers to a broader family of adaptation techniques that freeze most pretrained parameters and optimize only a small number of task-specific parameters. Prompt learning (\ref{sub2:Promptlearning}) is an important form of PEFT, while other representative techniques include adapters~\cite{hu2023llm}, LoRA, prefix-based modules~\cite{li2021prefix}, and bias-only tuning~\cite{zaken2022bitfit}.

PEFT is particularly suitable for federated LLM adaptation. Because only compact parameter subsets are transmitted, communication overhead can be reduced from full-model scale to prompt- or adapter-module scale. Clients can maintain a common frozen backbone while learning local adaptation modules, and the server can aggregate only the lightweight updates. 

\subsubsection{Federated Prompt Learning (FPL)} 

FPL integrates federated optimization with prompt learning or closely related PEFT methods~\cite{3692070.3692451}. It mainly federates lightweight adaptation modules rather than the complete foundation model. FPL can substantially reduce communication and client-side optimization costs because only compact modules are trained and exchanged. In FPL, the frozen foundation model preserves broadly transferable knowledge and reduces the risk of catastrophic forgetting during local adaptation. Meanwhile, modular prompts and adapters enable the coexistence of shared and personalized capabilities. A typical FPL training round contains four stages:
\begin{enumerate}
    \item The server distributes the current global prompt or adapter parameters to selected clients.
    \item Each client attaches the received module to a shared frozen foundation model and performs local optimization using its private data.
    \item Clients transmit only their updated prompt or adapter parameters to the server.
    \item The server aggregates the lightweight updates and constructs a new global module.
\end{enumerate}

Let $\phi$ denote the federated prompt or adapter parameters, while $\theta$ denotes the frozen backbone. The FPL objective can be represented as:
\begin{equation}
\min _\phi \sum_{k=1}^K \frac{n_k}{\sum_{j=1}^K n_j} F_k(\boldsymbol{\theta}, \boldsymbol{\phi}) .
\end{equation}
After local optimization, the server may perform weighted aggregation:
\begin{equation}
\phi^{t+1}=\sum_{k \in S_t} \frac{n_k}{\sum_{j \in S_t} n_j} \phi_k^{t+1} .
\end{equation}

Different from conventional FL, in which the communicated variable $w$ may contain all model parameters, FPL communicates only $\phi$, whose size is generally much smaller than that of the foundation model. For example, PROMPTFL~\cite{guo2023promptfl} replaces conventional full-model training with cooperative optimization of continuous prompts over a frozen foundation model, while FedPrompt~\cite{zhao2023fedprompt} extends soft prompt tuning to federated pretrained language models and supports prompt tuning, P-tuning~\cite{liu2022p}, and prefix tuning.

Nevertheless, FPL introduces challenges that differ from those of conventional FL. Prompt parameters learned by heterogeneous clients may encode semantically different concepts, making position-wise averaging ineffective. Data heterogeneity can therefore cause prompt misalignment, client drift, and negative transfer. A single global prompt may favor dominant clients and provide limited adaptation to minority domains. Personalized FPL addresses this issue by separating global knowledge from client-specific prompts or by generating customized prompts according to client updates.

\subsubsection{Privacy, Security, and Robustness Considerations}

Although keeping raw data local is an important privacy property of FPL, lightweight parameter exchange does not eliminate security risks. Prompt or adapter updates may still reveal information about local examples through gradient inversion, membership inference, or representation reconstruction. A malicious client may poison its local data or manipulate its prompt updates to insert a backdoor into the aggregated module. Conversely, a malicious or compromised server may distribute manipulated prompts, infer client properties, or exploit intermediate activations in split architectures.

\textbf{Answers to RQ1:
} FPL emerges from combining FL with prompt-based tuning of large pretrained (often vision-language) models, motivated primarily by the practical limitations of applying conventional FL to foundation models: high communication overhead, prohibitive on-device compute and memory demands, non-IID data heterogeneity across clients, and risks of overfitting or catastrophic forgetting under full-model fine-tuning. 
Rather than exchanging full model gradients or weights, FPL keeps the pretrained backbone frozen at both server and clients, and collaboratively learns only lightweight, task-specific prompts, such as continuous soft-prompt embeddings, prefix vectors, or prompt-generating networks.Then these prompts are aggregated via standard or adapted FL protocols (e.g., FedAvg, FedProx, personalized FL). 
This design is enabled by advances in parameter-efficient fine-tuning (e.g., prompt tuning, prefix-tuning, P-tuning v2) and foundation models such as CLIP, whose strong pretrained representations can be effectively steered through prompts alone.
Different from conventional FL, FPL reduces communication and computation by orders of magnitude, supports client-level personalization by decomposing prompts into shared and local components, and better preserves general pretrained knowledge. 
Compared to full-model federated fine-tuning, FPL avoids destructive weight-averaging under heterogeneous data, lowers system requirements by eliminating the need to store or transmit full gradients and optimizer states, and offers greater flexibility for supporting multiple tasks or domains without maintaining separate full model copies, positioning FPL as a parameter-efficient, communication-friendly paradigm for adapting foundation models in federated settings.

\section{Federated Learning for Large Language Models}

From the perspective of the interaction between FL and LLMs, this survey broadly categorizes the existing research into three types: (i) federated learning for large language models; (ii) large language model-enhanced federated learning; (iii) the synergistic integration of federated learning and large language models.

For the first category, federated learning is applied to address the challenges of training, adaptation, and deployment of large language models on distributed private data. In this relationship, the large language model is the primary object of learning and optimization, while FL provides a collaborative training mechanism that keeps data local. Based on the lifecycle of LLMs, this category can be further divided into federated pretraining, federated fine-tuning, federated instruction tuning, federated alignment, federated prompt learning, federated model compression, and federated continual learning, among other directions.

\subsection{Larger Language Models Optimization through Federated Learning}

Chen~\emph{et al}\cite{che2023federated} address the fundamental difficulty of applying FL to large language models LLMs, namely the prohibitive communication and computation costs incurred when updating massive model parameters under non-IID decentralized data. To overcome the limitations of existing prompt-based FL methods such as performance degradation, inefficient training, and client drift, the authors propose FedPepTAO by integrating parameter-efficient prompt tuning with a communication-efficient adaptive optimization strategy. FedPepTAO lies in a layer-importance scoring and lossless selection mechanism, which identifies and synchronizes only a subset of influential prompt layers while keeping other prompts locally updated, thereby significantly reducing communication overhead without sacrificing accuracy.  FedPepTAO is evaluated through extensive experiments on 10 NLP benchmarks using RoBERTa-Large and multiple decoder-based LLMs (e.g., GPT-2, LLaMA-3B, LLaMA-7B). The results demonstrate that FedPepTAO consistently achieves state-of-the-art accuracy up to 60.8\% improvement over baselines, while reducing training time by up to 97.59\%.

Raje~\cite{raje2024communication} systematically investigates communication-efficient training of LLMs in FL settings, addressing the dual bottlenecks of limited client-side computation and expensive wireless communication that hinder practical federated LLM deployment.
Considering that fine-tuning updates of pretrained LLMs are intrinsically low-rank, the work integrates LoRA-based parameter-efficient fine-tuning with communication-only sparsification, proposing federated LoRA with simple sparsity (FLoSS). Unlike prior approaches that prune adapters during training, FLoSS applies unstructured top-k sparsity exclusively during download and upload phases, while preserving dense local optimization to maintain model utility.

To address the largely overlooked limitation of existing federated instruction tuning methods for large language models, Qin~\emph{et al.}~\cite{qin2025federated} propose FedHDS.   Considering the resource-constrained edge devices and overfitting to narrow client-specific domains, FedHDS constructs a hierarchical selection strategy that combines cross-layer feature fusion from multiple transformer layers with density-based clustering, enabling both local redundancy removal and global coordination across clients in a privacy-preserving manner. FedHDS establishes the first systematic solution for federated data-efficient instruction tuning and highlights the importance of redundancy-aware data selection for scalable, efficient, and generalizable federated LLM training.

Furthermore, FedDQC~\cite{du2025feddqc} tackles a fundamental yet underexplored challenge in federated instruction tuning of large language models: data quality heterogeneity across decentralized clients, which can severely degrade global model performance despite privacy preservation. 
To address the lack of global visibility and the impracticality of centralized data filtering in FL, the authors~\cite{du2025feddqc} propose FedDQC. It is a novel federated data quality control framework that operates entirely on the client side with minimal overhead. 
The kernel of FedDQC is the instruction response alignment (IRA) metric, a lightweight and privacy-preserving quality estimator that measures how well an instruction conditions its response by comparing conditional and unconditional inference losses. Building on IRA, FedDQC introduces a quality-aware hierarchical federated training strategy that progressively fine-tunes the model from high-IRA (easy, high-quality) samples to lower-IRA (harder, noisier) data, mirroring human curriculum learning. 

To address the data heterogeneity issue in federated prompt tuning, Chen~\emph{et al.}\cite{chen2025dualfpt} propose DualFPT, a federated visual prompt-tuning framework that jointly pursues generalized and personalized adaptation by decomposing learnable prompts into shared global prompts and client-specific local prompts. Because feature shifts and class imbalance impair the transferability of globally aggregated prompts while purely personalized methods risk overfitting to client-specific distributions. DualFPT is compose of two mechanisms: (i) feature sharing (FS); (ii) prompt composition scheme (PCS). First, FS uses a variational autoencoder to disentangle classification-sensitive features from redundant information and applies differential-privacy noise before sharing selected features, narrowing inter-client distribution gaps to improve global-prompt generalization. Second, the PCS employs a lightweight adaptive network to estimate similarity between each test instance and client distributions, dynamically combining local prompts into an instance-specific composite prompt. A two-stage alternating strategy optimizes global parameters, local prompts, and the adaptive network.
Nevertheless, DualFPT requires an added feature-extraction stage, exchange of differentially private intermediate features, and training of auxiliary VAE, classifier, and adaptive-network components, increasing implementation complexity relative to prompt-only aggregation. Evaluation is also limited to image classification with a ViT-B/16 backbone, controlled feature- and class-heterogeneity settings, fixed prompt configurations, and relatively short federated training schedules.

To address the task and client resource heterogeneity in cross-device FL, Bai~\emph{et al.}~\cite{bai2024federated} propose FlexLoRA, a federated fine-tuning framework for LLMs. Motivated by the ``bucket effect" in conventional FL where all clients are constrained by the least-capable participant, the authors allow clients to apply heterogeneous LoRA ranks to match their local resources, enabling more powerful clients to contribute richer, less task-specific knowledge. FlexLoRA aggregates client updates by reconstructing full LoRA weight matrices, averaging them on the server, and applying SVD-based decomposition to redistribute rank-adaptive LoRA parameters back to clients, without additional hyperparameters. Theoretical analysis links higher local ranks and larger client populations to improved generalization bounds.

Gao \emph{et al.}~\cite{gao2025federated} address the memory overhead and training latency of federated LoRA fine-tuning for LLMs on resource-constrained, system-heterogeneous edge devices by proposing FAH-QLoRA, a framework that combines heterogeneous base-model quantization with dynamically adjusted LoRA ranks. Its core mechanism follows a two-stage rank-allocation strategy: first determining the average LoRA rank that maximizes loss-reduction rate per unit wall-clock time, then assigning device-specific ranks based on heterogeneous computation and communication capabilities, allocating lower ranks to slower devices to mitigate the straggler effect. Truncation and zero-padding further support local training and global aggregation of heterogeneous LoRA modules. The main contribution lies in jointly optimizing model precision and LoRA rank allocation to improve time and memory efficiency in federated LLM fine-tuning, complemented by a convergence analysis under non-convex, non-IID settings.

\subsection{Larger Language Models Application via Federated Learning}
FedMRG~\cite{che2025llm} is the first comprehensive framework for LLM-driven medical report generation (MRG) under federated learning, tackling two key obstacles to multi-center collaboration: prohibitive LLM communication costs and severe multi-modal data heterogeneity across institutions. To enable privacy-preserving, scalable training, FedMRG employs LoRA to substantially reduce communication overhead. It further addresses heterogeneity at both visual and textual levels: on the encoder side, Hierarchical Contrasting and Prompting (HCP) combines client-aware contrastive learning with diagnosis-aware prompting to capture globally consistent yet locally distinctive visual features; on the decoder side, a Dual-adapter Mutual Boosting (DMB) mechanism harmonizes global reporting knowledge with client-specific linguistic styles via bidirectional knowledge distillation.

Otoum~\emph{et al.}~\cite{otoum2025llms} present an LLM-driven federated learning framework for scalable and secure IoT management, motivated by the latency, privacy, energy, and scalability limitations of centralized cloud-based IoT architectures. The authors integrate LLMs with FL in a hybrid edge–cloud architecture, enabling privacy-preserving decision-making directly on IoT devices while leveraging cloud resources for global coordination. A key contribution is the gradient sensing federated strategy (GSFS), which adaptively regulates client participation and asynchronous update uploads based on performance shifts and gradient magnitudes, reducing redundant communication and improving convergence efficiency over classical methods, such as FedAvg and FedOpt.

Evaluations on the IoT-23 dataset show GSFS achieves higher accuracy and F1-scores for both central and client models, while significantly cutting response latency (up to 51\% on the client side) and improving energy efficiency. Results further indicate that edge-based LLM inference enables real-time IoT analytics with reduced cloud dependency, while federated coordination preserves data privacy across heterogeneous devices.

Agarwal et al.~\cite{agarwal2023practical} provide a critical empirical investigation into whether FL with pretrained language models (PLMs) genuinely achieves domain adaptation and personalization, or whether its apparent success stems largely from pretraining-induced regularization. Through systematic experiments on three NLP tasks: sentiment classification (SST-2), sequence tagging (OntoNotes), and text generation (Gigaword), the authors analyze key confounders in federated NLP, including the role of pretrained weights, client size imbalance, data partition strategies, and the trade-off between server generalization and client personalization.
Results show that with PLMs, the performance gap between federated and centralized training stays small even after aggressively ablating client updates, suggesting FL often fails to meaningfully adapt to local client distributions and instead benefits from pretraining's strong semantic priors. The proposed personalization–generalization slope ($m\Delta P$) further indicates that local-data learning provides limited or neutral gains to global generalization, challenging common claims about personalization in federated NLP. While uniform client data distributions improve convergence speed and stability, they do not fundamentally resolve the lack of true domain adaptation.

Zhang~\cite{zhang2025fed} addresses the resource and memory heterogeneity encountered in federated foundation-model fine-tuning, where resource-constrained clients may be unable to update all LoRA layers efficiently. It proposes Fed-HeLLo, a federated LoRA framework that assigns different subsets of trainable LoRA layers to clients according to their computational capabilities and layer importance. Its core design combines Fisher information matrix–based allocation (FIM-HLA) for dynamically estimating layer importance with geometrically defined and randomized allocation strategies (GD-HLA/RGD-HLA) for stabilizing early-stage training. The principal contribution therefore lies in jointly exploiting client resource heterogeneity and layer-wise importance for efficient federated LoRA fine-tuning. Nevertheless, the framework still relies on a server-side proxy dataset, retains non-negligible activation-memory requirements, lacks explicit privacy protection, and does not directly resolve data heterogeneity.

\subsection{Empowering the Reasoning Process of LLMs in Federated Settings}

eFedLLM~\cite{ding2024efedllm} is an FL-based framework for LLM inference, targeting the prohibitive computational, memory, and bandwidth requirements that limit LLM accessibility. Departing from conventional data-parallel FL, it adopts a transformer-based model-parallel architecture, distributing different transformer layers across heterogeneous participants to enable collaborative inference without requiring any single user to host the full model.
To ensure reliability in this chained execution setting, eFedLLM introduces a trust-based incentive and verification mechanism that evaluates intermediate layer outputs and filters malicious or low-quality contributors. It further incorporates transformer-tailored optimizations, including hierarchical memory access strategies to reduce global memory reads and SVD-based low-rank compression of weight matrices to lower communication bandwidth while preserving accuracy. Analytical and numerical evaluations show these optimizations reduce memory access and bandwidth usage by up to 60\% under practical compression ratios, significantly improving inference efficiency.

Lin~\cite{lin2024splitlora} introduces SplitLoRA, the first split learning-based parameter-efficient fine-tuning framework for LLMs, addressing the prohibitive computation and communication costs limiting federated LLM adaptation on distributed private data. Motivated by the scarcity of high-quality public data and the impracticality of full-model federated fine-tuning, SplitLoRA combines split federated learning (SFL) with LoRA-based PEFT, partitioning the LLM between clients and a central server so clients train only shallow layers with lightweight LoRA adapters while the server handles most computation. This substantially reduces client-side resource demands and mitigates data heterogeneity by centralizing deeper representations.
SplitLoRA provides a practical balance between performance, efficiency, and scalability. Meanwhile, it establishes an open-source benchmark for split LLM fine-tuning. However, open challenges remain in optimal model splitting, handling heterogeneous client resources, and strengthening privacy guarantees against split-learning inference attacks, pointing to promising directions for future research.

\begingroup
\renewcommand{\arraystretch}{1.18}
\setlength{\tabcolsep}{4pt}

\begin{xltabular}{\textwidth}{
    >{\raggedright\arraybackslash}p{0.17\textwidth}
    >{\raggedright\arraybackslash}p{0.15\textwidth}
    >{\raggedright\arraybackslash}X
    >{\raggedright\arraybackslash}X
}
\caption{Research trajectories of FL for LLMs.}\label{tab:fedllm_research_trajectories}\\
\toprule
\textbf{Research Trajectory} &
\textbf{Representative Works} &
\textbf{Core Idea} &
\textbf{Main Challenges} \\
\midrule
\endfirsthead

\multicolumn{4}{c}{\tablename~\thetable\ (continued)}\\
\toprule
\textbf{Research Trajectory} &
\textbf{Representative Works} &
\textbf{Core Idea} &
\textbf{Main Challenges} \\
\midrule
\endhead

\midrule
\multicolumn{4}{r}{Continued on next page}\\
\endfoot

\bottomrule
\endlastfoot

\textbf{Parameter-Efficient Adaptation and Communication Compression}
&
FedPepTAO~\cite{che2023federated}, FLoSS~\cite{raje2024communication}
&
Replace full-model federated fine-tuning with lightweight prompt- or LoRA-based adaptation.
&
How to minimize communication overhead while preserving the effectiveness and stability of local optimization.
\\

\midrule

\textbf{Federated Instruction Data Selection and Quality Control}
&
FedHDS~\cite{qin2025federated}, FedDQC~\cite{du2025feddqc}
&
Hierarchical data selection, cross-layer feature fusion, density-based clustering, and quality-aware curriculum learning.
&
How to identify which decentralized training samples are informative and reliable when raw client data cannot be globally inspected.
\\

\midrule

\textbf{Data Heterogeneity and Personalization}
&
DualFPT~\cite{chen2025dualfpt}, Agarwal et al. \cite{agarwal2023practical}
&
Address non-IID client distributions through global--local prompt decomposition and personalized adaptation.
&
How to balance globally transferable knowledge and client-specific adaptation without sacrificing either generalization or personalization.
\\

\midrule

\textbf{System and Resource Heterogeneity}
&
FlexLoRA~\cite{bai2024federated}, FAH-QLoRA \cite{gao2025federated}, Fed-HeLLo~\cite{zhang2025fed}
&
Allow heterogeneous clients to train different LoRA ranks, quantization precisions, or subsets of LoRA layers according to devices' capabilities.
&
How to prevent resource-constrained clients from becoming system bottlenecks while fully exploiting the clients with high capabilities.
\\

\midrule

\textbf{Model Partitioning and Collaborative Inference}
&
eFedLLM~\cite{ding2024efedllm}, SplitLoRA \cite{lin2024splitlora}
&
Partition LLMs across clients, servers, or multiple participants so that no single resource-constrained device is required to host or execute the full model.
&
How to achieve efficient and trustworthy collaborative execution while controlling intermediate-feature leakage, communication latency, participant heterogeneity, and model-partitioning overhead.
\\

\midrule

\textbf{Application-Driven Integrated Optimization}
&
FedMRG~\cite{che2025llm}, GSFS~\cite{otoum2025llms}
&
Integrate parameter-efficient adaptation, heterogeneity handling, communication optimization, and edge--cloud collaboration according to the requirements of specific domains, such as healthcare and IoT.
&
How to adapt general-purpose federated LLM techniques to domain-specific data distributions, system constraints, privacy requirements, and real-time deployment conditions.
\\

\end{xltabular}
\endgroup

As summarized in Table~\ref{tab:fedllm_research_trajectories}, the research focus of federated LLMs has gradually shifted from simply reducing the cost of full-model federated optimization toward a more fine-grained design space involving \emph{what to train},
\emph{what to communicate}, \emph{where to execute}, and \emph{what knowledge should be shared}.
Early efforts primarily exploit parameter-efficient adaptation and communication compression, whereas more recent studies increasingly consider data quality, client heterogeneity, and collaborative model execution in a unified manner. This evolution suggests that federated LLM optimization is no longer merely an aggregation problem, but rather a joint optimization problem
across model parameters, data, communication, computation, and privacy.

\section{Large Language Model-enhanced Federated Learning}

In LLM-enhanced FL, LLMs serve as knowledge auxiliary tools to improve the performance and automation level of conventional federated learning. In this context, the model being federated is not necessarily a large language model itself, but may instead be a classification model, a vision model, or another lightweight model. 
LLMs can participate in local data generation, data augmentation, automatic annotation, and pseudo-label construction to alleviate issues of insufficient client data and class imbalance.
They can also transfer knowledge to federated models through knowledge distillation, feature transfer, and model initialization. Furthermore, LLMs can assist with client selection, aggregation strategy design, hyperparameter configuration, training log analysis, anomalous update detection, and privacy risk interpretation, thereby improving the optimization efficiency, manageability, and intelligence level of federated learning systems.

As summarized in Table~\ref{tab:llm_enhanced_fl_trajectories}, the role of large pretrained models in federated learning is gradually evolving from being directly federated toward serving as reusable knowledge priors for lightweight collaborative adaptation. Accordingly, the main optimization target shifts from complete model parameters to prompts, adapters, low-rank modules, and compressed updates.

This evolution further relaxes several conventional assumptions of
federated learning. In particular, clients are no longer necessarily required to optimize the same parameter space or even possess the full foundation model. Instead, pretrained knowledge can be accessed through
parameter-efficient adaptation, off-site tuning, lightweight emulators, or domain-specific instruction tuning. Consequently, LLM-enhanced federated learning increasingly becomes a joint design problem involving knowledge reuse, parameter efficiency, model accessibility, data quality, and communication efficiency.
\begingroup
\renewcommand{\arraystretch}{1.18}
\setlength{\tabcolsep}{4pt}

\begin{xltabular}{\textwidth}{
    >{\raggedright\arraybackslash}p{0.15\textwidth}
    >{\raggedright\arraybackslash}p{0.15\textwidth}
    >{\raggedright\arraybackslash}X
    >{\raggedright\arraybackslash}X
}
\caption{Research trajectories of large language model-enhanced federated learning.}\label{tab:llm_enhanced_fl_trajectories}\\
\toprule
\textbf{Research Trajectory} &
\textbf{Representative Works} &
\textbf{Core Idea} &
\textbf{Main Challenges} \\
\midrule
\endfirsthead

\multicolumn{4}{c}{\tablename~\thetable\ (continued)}\\
\toprule
\textbf{Research Trajectory} &
\textbf{Representative Works} &
\textbf{Core Idea} &
\textbf{Main Challenges} \\
\midrule
\endhead

\midrule
\multicolumn{4}{r}{Continued on next page}\\
\endfoot

\bottomrule
\endlastfoot

\textbf{Foundation Model-Assisted Federated Adaptation}
&
PROMPTFL~\cite{guo2023promptfl}, FedPrompt~\cite{zhao2023fedprompt}
&
Exploit the transferable knowledge encoded in pretrained foundation models while freezing the backbone and collaboratively optimizing only lightweight prompt parameters. 
&
How to leverage strong pretrained knowledge to reduce communication
and optimization costs while retaining sufficient adaptability to
heterogeneous and highly specialized downstream tasks.
\\

\midrule

\textbf{PEFT-Based Practical Federated Learning}
&
FederatedScope-LLM~\cite{kuang2024federatedscope},
Malaviya et al. \cite{malaviya2023reducing}
&
Apply parameter-efficient fine-tuning techniques, including LoRA~\cite{hu2022lora},
Prefix-Tuning, Adapters, BitFit~\cite{zaken2022bitfit}, and Prompt-Tuning, to reduce the trainable and communicated parameter space. &
How to select and configure PEFT mechanisms that provide a favorable
trade-off among model utility, communication efficiency, client-side
resource consumption, and robustness to non-IID data.
\\

\midrule

\textbf{Model-Access-Constrained and Off-Site Federated Tuning}
&
FedBiOT \cite{wu2024fedbiot}
&
Relax the assumption that each client must access or deploy the full
foundation model. The original model is represented through lightweight
emulator and adapter components, while client-side adapter optimization
is coordinated with server-side emulator distillation through bi-level
optimization.
&
How to simultaneously protect model intellectual property, reduce
client-side computation and memory requirements, and preserve
adaptation quality under distribution mismatch between server-side
public data and client-side private data.
\\

\midrule

\textbf{Data- and Application-Aware LLM-Assisted Federated Learning}
&
FIT with Feature Diversity~\cite{chen2024empowering}
&
Combine parameter-efficient federated instruction tuning with systematic augmentation of instruction-following data. Foundation-model knowledge and feature diversity are exploited to compensate for scarce
annotations.
&
How to exploit pretrained knowledge and diversified training data to
improve downstream generalization while preserving privacy and avoiding
excessive dependence on application-specific data augmentation
strategies.
\\

\midrule

\textbf{Communication-Aware Lightweight Adaptation}
&
FLM-TopK~\cite{qiu2025flm}
&
Further compress PEFT updates through joint gradient sparsification and
quantization. Intervalized TopK sparsification reduces not only the
number of transmitted gradient values but also the position-index
overhead associated with conventional TopK compression.
&
How to further reduce communication after the trainable parameter space
has already been substantially reduced by PEFT, while controlling the
optimization errors introduced by sparsification and quantization.
\\

\end{xltabular}
\endgroup

\subsection{Using LLMs to Empower Federated Learning Settings}

Motivated by the severe communication overhead, slow convergence, data scarcity, and non-IID sensitivity of classical FL, Guo \emph{et al.}~\cite{guo2023promptfl} leverage the strong generalization capability of large pretrained vision–language models (e.g., CLIP) and
propose a paradigm shift in FL, named PROMPTFL.
PROMPTFL allows distributed clients to collaboratively optimize only a small set of continuous (soft) prompt parameters on top of a frozen foundation model instead of conventional full-model training. This design dramatically reduces trainable parameters and communication cost while preserving privacy, as clients share only prompt updates rather than raw data or full model gradients. Theoretically, PROMPTFL can reach a convergence rate of $\mathcal{O}(1/\sqrt{T})$ under the setting where partial clients are selected at each round, and local data distributions are Non-IID.
Extensive experiments across diverse vision benchmarks demonstrate that PROMPTFL achieves competitive or superior accuracy and robustness. Particularly in heterogeneous and data-scarce scenarios, it uses orders of magnitude fewer parameters and fewer communication rounds than FedAvg and fine-tuning-based baselines. Its performance depends on the availability of strong foundation models and may be less effective for highly specialized tasks.

FedPrompt~\cite{zhao2023fedprompt} extends parameter-efficient learning to the federated setting by integrating soft prompt tuning with federated learning for large pre-trained language models (PLMs). Considering the prohibitive communication and memory costs of full-model fine-tuning in FL, FedPrompt freezes the PLM backbone and collaboratively trains only a small set of soft prompt parameters using a split-aggregation strategy, reducing communication overhead to approximately 0.01\% of the original model size with minimal accuracy degradation. The method is compatible with standard FedAvg and supports multiple prompt paradigms (e.g., prompt tuning, P-tuning, prefix-tuning). Extensive experiments on diverse NLP tasks under both IID and Non-IID data distributions demonstrate that FedPrompt achieves competitive performance compared to full fine-tuning while being significantly more communication-efficient. Furthermore, robustness analyses show that FedPrompt is resistant to backdoor attacks based on data poisoning, as prompt aggregation mitigates malicious updates, and optional LDP can be incorporated to further enhance privacy at the cost of moderate accuracy loss. 

\subsection{Prompt Adapter for Federated Learning}

FederatedScope-LLM (FS-LLM) \cite{kuang2024federatedscope} is a comprehensive, open-source framework designed to bridge the gap between general-purpose FL systems and the unique requirements of federated fine-tuning of LLMs. Motivated by the prohibitive communication cost, computation burden, data heterogeneity, and model intellectual-property constraints inherent to LLMs, FS-LLM provides an end-to-end solution that integrates benchmarking, algorithm support, and system-level optimization. It offers a standardized federated LLM benchmarking pipeline with curated datasets and evaluation tasks, enabling fair comparison across domains such as code generation, general language understanding, and reasoning. To address efficiency, FS-LLM incorporates a rich suite of parameter-efficient fine-tuning (PEFT) methods (e.g., LoRA, P-/prompt-tuning) and supports federated offsite tuning for scenarios where clients cannot access full models, achieving orders-of-magnitude reductions in communication while maintaining competitive performance. At the system level, FS-LLM integrates accelerating and resource-efficient operators (mixed precision, compression, DeepSpeed, offloading) and flexible training modes, making federated LLM fine-tuning feasible on limited hardware.

Malaviya~\emph{et al.}~\cite{malaviya2023reducing} investigate how parameter-efficient fine-tuning (PEFT) can make FL with pretrained language models (PLMs) practical under strict privacy constraints and limited communication budgets. Motivated by the prohibitive cost of transmitting full PLM parameters and the frequent lack of labeled client data, the authors systematically evaluate four PEFT methods: Prefix-tuning, Adapters, BitFit, and LoRA, under both supervised FedAvg and a more realistic semi-supervised scenario (FedFAME). Only a small subset of trainable ``$\delta$" parameters is exchanged between clients and server while the PLM backbone stays frozen, reducing communication from about 420 MB to below 50 MB, and in some cases under 1 MB.
Experiments on six GLUE tasks under varying non-IID conditions show PEFT methods achieve competitive performance, with Prefix and Adapter tuning generally outperforming BitFit and LoRA, and FedFAME exhibiting substantially higher robustness than FedAvg under severe heterogeneity ($\alpha$ = 0.1). The paper further demonstrates strong task-level transferability of PEFT parameters, particularly between semantically related tasks, enabling effective zero-shot adaptation when target-task labels are scarce. While the approach assumes the frozen PLM fits on client devices and is sensitive to hyperparameters like learning rate, the work provides compelling evidence that PEFT is a viable, scalable foundation for communication-efficient, privacy-preserving federated NLP.

To relieve the two critical bottlenecks in federated LLM fine-tuning: clients' lack of access to full model parameters and prohibitive computation/communication costs, Wu et al.~\cite{wu2024fedbiot} propose FedBiOT, a resource-efficient, privacy-preserving framework that avoids full-model exposure. 
Building on the off-site-tuning paradigm, FedBiOT compresses the LLM into an emulator, aligned server-side on public data to mimic the original model, and an adapter, which captures domain-specific knowledge and is fine-tuned locally by clients using lightweight LoRA modules. To handle the distribution mismatch between server-side public data and client-side private data, the authors formulate a bi-level optimization problem alternating between client-side adapter updates and server-side emulator distillation.
Experiments on LLaMA-2 across math problem solving, code generation, and question answering under both IID and non-IID settings show FedBiOT consistently outperforms offsite-tuning and FedOT, achieving higher accuracy and robustness while cutting client-side computation by up to 50\% and avoiding transmission of full model parameters. Though it incurs slightly higher communication overhead than prior offsite methods due to emulator updates, this cost remains negligible compared to full LLM transfer.

Chen \emph{et al}~\cite{chen2024empowering} propose federated instruction tuning (FIT) with a complementary feature diversity (FD) strategy to enable privacy-preserving and data-efficient deployment of large language models for IoT-based autonomous driving. Motivated by the scarcity of high-quality annotated driving data and stringent privacy constraints in vehicular IoT environments, FIT leverages federated learning to collaboratively fine-tune only lightweight adapter parameters of a frozen multimodal LLM, thereby substantially reducing communication and computational overhead while preserving sensitive local data. To further address limited scenario coverage and improve generalization, the FD strategy systematically augments instruction-following data along key dimensions, including time, weather, and occlusion, enriching both visual and textual diversity and enhancing robustness in open-world driving conditions. Extensive experiments on the DriveLM benchmark using LLaMA-Adapter backbones and multiple FL algorithms demonstrate that FIT combined with FD, consistently outperforms local training and achieves strong gains in accuracy, language quality, and semantic alignment, while synchronizing only a tiny fraction of model parameters.

\subsection{Prompt Tuning for Federated Learning}

Qiu~\emph{et al.}~\cite{qiu2025flm} solve the substantial communication overhead incurred when exchanging high-dimensional LoRA updates during federated large language model fine-tuning, particularly because conventional TopK sparsification requires transmitting position identifiers for every retained gradient. To this end, the authors propose FLM-TopK, a communication-efficient framework that partitions gradients into multiple intervals and independently applies TopK sparsification within each interval, thereby reducing the number of bits required to encode gradient positions.
Its core design models both original and intervalized fine-tuning gradients using a zero-mean Gaussian distribution and formulates compression as a joint optimization of interval size, packet allocation, and interval-specific sparsification rates under a fixed communication budget. Because the resulting optimization problem is non-convex, the authors decompose it into alternating convex subproblems and combine sparsification with unbiased quantization to balance discarded-gradient and quantization errors. Nevertheless, the formulation relies on an empirically derived Gaussian-gradient assumption and a fixed per-round communication budget, while the evaluation adopts a single LoRA rank, simulated participation of 10 out of 100 clients, and limited network-level measurements.

\section{The Synergistic Integration of FL and LLMs}

The synergistic integration of FL and LLMs no longer exhibits a one-directional service relationship but instead jointly forms a distributed intelligent system with bidirectional interaction capabilities. 
As summarized in Table~\ref{tab:synergistic_fl_llm_trajectories}, the synergistic integration of federated learning and large language models extends collaboration beyond conventional model-parameter aggregation. The objects of federation progressively evolve from shared model components toward personalized prompts, prompt mixtures, cross-task adapters, semantic representations, and black-box optimization statistics.

This evolution also relaxes several assumptions commonly adopted in traditional federated learning. Participants are no longer necessarily required to maintain identical model parameterizations, perform the same downstream task, or possess white-box access to the underlying foundation model. Instead, collaboration can occur at the level of adaptation modules, semantic concepts, multimodal task knowledge, or prompt evaluation information.
Consequently, synergistic FL--LLM systems exhibit a broader transition from \emph{parameter-level federation} toward \emph{knowledge-level federation}, where the central problem is no longer merely how to aggregate distributed updates, but how to identify, align, personalize, and securely transfer useful knowledge across heterogeneous participants.

\begingroup
\small
\renewcommand{\arraystretch}{1.05}
\setlength{\tabcolsep}{3pt}

\begin{xltabular}{\textwidth}{
    >{\raggedright\arraybackslash}p{0.15\textwidth}
    >{\raggedright\arraybackslash}p{0.15\textwidth}
    >{\raggedright\arraybackslash}X
    >{\raggedright\arraybackslash}X
}
\caption{Research trajectories of the synergistic integration of FL and LLMs.}\label{tab:synergistic_fl_llm_trajectories}\\
\toprule
\textbf{Research Trajectory} &
\textbf{Representative Works} &
\textbf{Core Idea} &
\textbf{Main Challenges} \\
\midrule
\endfirsthead

\multicolumn{4}{c}{\tablename~\thetable\ (continued)}\\
\toprule
\textbf{Research Trajectory} &
\textbf{Representative Works} &
\textbf{Core Idea} &
\textbf{Main Challenges} \\
\midrule
\endhead

\midrule
\multicolumn{4}{r}{Continued on next page}\\
\endfoot

\bottomrule
\endlastfoot

\textbf{Personalized Foundation-Model Adaptation}
&
FedPerfix~\cite{sun2023fedperfix}, pFedPG~\cite{yang2023efficient}
&

Personalization can further be generated from client updates through a server-side prompt generator instead of relying on a single globally aggregated adaptation module.
&
How to retain transferable knowledge from the shared foundation model while providing communication-efficient personalization for clients with severely non-IID data, limited local samples, and heterogeneous resource constraints.
\\

\midrule

\textbf{Mixture-Based and Global--Local Prompt Personalization}
&
pFedMoAP~\cite{luo2025mixture}, FedPGP~\cite{3692070.3692451}
&
Model personalized adaptation either as a mixture of local and
non-local prompt experts or as a decomposition of a globally shared prompt and a locally retained low-rank component. &
How to benefit from cross-client knowledge transfer without sacrificing local specialization or degrading the generalization capability inherited from pretrained foundation models.
\\

\midrule

\textbf{Domain-Generalized and Privacy-Aware Prompt Learning}
&
PLAN~\cite{gong2026federated}, DP-FPL~\cite{tran2025privacy}
&
Use lightweight prompts as a medium for transferring domain knowledge across clients and further integrate personalized prompt adaptation with low-rank decomposition and DP to support unseen-domain generalization and formal privacy protection.
&
How to jointly balance personalization, cross-domain generalization, and privacy, since aggressive local adaptation may impair transferable knowledge while stronger privacy protection may introduce substantial utility degradation.
\\

\midrule

\textbf{Multimodal and Task-Heterogeneous Federated Adaptation}
&
Pilot~\cite{xiong2025pilot}
&
Task- and client-specific adapters extract specialized knowledge, while cross-task adapter mixtures and task-aware aggregation promote selective knowledge transfer.
&
How to enable beneficial knowledge sharing among clients performing different multimodal tasks while suppressing negative transfer caused by task, modality, and data heterogeneity.
\\

\midrule

\textbf{Semantic and Robustness-Oriented Prompt Aggregation}
&
PFPT~\cite{weng2024probabilistic}, Wang et al.~\cite{Wang2025empirical}, FOCoOp~\cite{Liao2025FOCoOp}
&
Move beyond position-wise parameter averaging by aligning local prompts at the semantic or concept level, systematically characterizing aggregation behavior under heterogeneous client distributions, and explicitly optimizing both in-distribution performance and out-of-distribution robustness.
&
How to construct globally useful prompt representations when local prompts may be semantically misaligned, client distributions differ, domain shifts occur, and previously unseen distributions must be handled at inference time.
\\

\midrule

\textbf{Black-Box Collaborative Prompt Optimization}
&
FedPOB~\cite{lu2025fedpob}
&
Bandit optimization\cite{agarwal2011stochastic}, preference feedback, summary-statistic aggregation, and event-triggered communication are used to coordinate black-box prompt
search.
&
How to achieve query-efficient, communication-efficient, and
privacy-preserving collaborative prompt optimization when the underlying foundation model is proprietary and only black-box access and limited feedback are available.
\\
\midrule

\textbf{Privacy, Security, and Trusted Execution}
&
FL-GLM~\cite{zheng2024safely}, Huang et al.~\cite{huang2024fast}, ROFED-LLM \cite{wang2025rofed}
&
Combine split learning, encryption, TEEs, differential privacy (DP), secure aggregation, model pruning, and communication-layer defenses to protect federated LLM training against information leakage and adversarial attacks.
&
How to strengthen privacy and robustness without introducing prohibitive computation, communication, or system complexity.
\\

\end{xltabular}
\endgroup

\subsection{Federated pruning learning for personalized model learning}

FedPerfix~\cite{sun2023fedperfix} advances personalized federated learning (PFL) for vision transformers by systematically addressing where and how to personalize ViT models under heterogeneous data. Through an empirical sensitivity analysis, the authors identify the self-attention layers and classification head as the most distribution-sensitive components of ViTs, motivating partial personalization rather than full local training. Building on this insight and drawing an analogy to transfer learning, FedPerfix introduces prefix-based personalization with parallel attention, where lightweight, client-specific prefix plugins adapt the global self-attention representations while the backbone remains globally aggregated. Yang et al.~\cite{yang2023efficient} propose pFedPG, a personalized federated learning framework that uses client-specific prompt generation to efficiently adapt large pretrained foundation models under severe data heterogeneity, communication constraints, and limited client resources. Instead of sharing or averaging full model parameters, pFedPG freezes the backbone (e.g., ViT) and alternates between local personalized prompt adaptation and server-side personalized prompt generation, where a lightweight cross-attention-based generator infers client-specific optimization directions from prompt updates and produces customized prompts for each client. This design addresses key challenges of federated learning with foundation models—performance degradation under non-IID data, excessive communication cost, and overfitting with limited local data—while retaining strong representational power.

For the personalization issues under severe data heterogeneity in federated prompt learning, Luo \emph{et al.} \cite{luo2025mixture} argue that distributing only a single globally aggregated prompt fails to fully exploit the lightweight nature of prompt parameters. Hence, the authors propose personalized federated mixture of adaptive prompts (pFedMoAP) that reformulates locally trained prompts as specialized experts and lets clients download multiple pre-aggregated prompts from other participants as fixed non-local experts. 
The design maintains a server-side prompt pool and uses prompt-space K-nearest-neighbor selection to assign relevant non-local experts to each client; a client-specific attention-based gating network then takes image features as queries and local/non-local text features as keys and values to produce instance-adaptive text representations. Final predictions combine the mixture-enhanced representation with the locally trained prompt, while only the local prompt is uploaded and the gating network stays private.
Ablations show jointly retaining local and mixture-based logits improves performance, moderate local-prompt weighting generally suffices, and more experts help but with diminishing returns. Gating-feature dimensions of 128–256 balance representation capacity and parameter efficiency. However, the method requires a server-side expert pool, multiple prompt downloads, and a client-specific gating network. Thus, communication, storage, and computation shall grow with expert count despite prompts being lightweight.

In federated prompt learning, the personalization and generalization are hard to balance. Because a shared prompt may inadequately represent heterogeneous client distributions while aggressive local adaptation erodes the transferable knowledge of pretrained vision-language models. To this end, Cui~\emph{et al.}~\cite{3692070.3692451} design FedPGP, a framework that preserves CLIP's generalization while adapting the global prompt to client-specific data, representing each personalized prompt as the sum of an aggregated global prompt and a locally retained low-rank adaptation term. FedPGP further introduces a prompt-wise contrastive objective treating global and handcrafted CLIP prompts as positive pairs to preserve category-agnostic knowledge, while treating global and personalized prompts as negative pairs to encourage client-specific representations; only the global prompt is communicated and aggregated, while low-rank adaptation parameters remain local. Ablations confirm this design: full-rank adaptation improves local accuracy but substantially degrades base-to-novel generalization, while removing either contrastive relation reduces overall performance. Nevertheless, the framework is evaluated only on CLIP-based image classification, relies on fixed prompt and bottleneck dimensions, and lacks a theoretical explanation for why low-rank adaptation achieves the observed generalization–personalization trade-off.

Federated domain generalization means that models trained collaboratively on heterogeneous source domains must generalize to unseen domains without exposing sensitive client data. The conventional approaches exchange domain knowledge directly derived from local samples such as feature statistics, class prototypes, or frequency information, which might bring privacy risks. Therefore, Gong~\emph{et al.}~\cite{gong2026federated} propose \textbf{P}rompt \textbf{L}earning and \textbf{A}ggregatio\textbf{N} (PLAN), which utilizes locally learned prompts as a parameter-efficient and potentially less data-revealing medium for cross-client knowledge transfer. PLAN lies in combining reference-guided local prompt learning with attention-based multimodal prompt aggregation to improve unseen-domain generalization without exchanging raw data or directly extracted domain statistics. It follows a two-stage procedure each round. First, every client learns multi-layer text and visual prompts on local data while aligning predictions with the previous round's global prompts via a KL-divergence regularizer, indirectly synchronizing client prompts and reducing local-domain overfitting. Second, lightweight attention-based aggregators estimate the relative importance of local prompts and selectively combine them into global text and visual prompts, rather than relying on fixed or uniform averaging. Only prompt tokens and aggregator parameters are optimized, while the pretrained CLIP backbone stays frozen.
However, the privacy advantage rests mainly on the indirect nature of prompt learning rather than formal guarantees or leakage analysis. Distributing all local prompts to clients and performing two communication phases per round may also constrain scalability as participant numbers grow, and evaluation is restricted to CLIP-based image classification with a shared label space between source and target domains.

While Tran~\emph{et al.}~\cite{tran2025privacy} investigate the competing requirements of personalization, generalization, and privacy in FPL for multimodal LLMs. Because personalized prompts can overfit heterogeneous local data and generalize poorly to unseen inputs, while differentially private training may substantially reduce both local and cross-domain performance. Thus, Tran~\emph{et al.} proposes DP-FPL, a differentially private personalized federated prompt-learning framework designed to protect prompts against membership-inference attacks while preserving their utility. In DP-FPL, each personalized prompt combines the global prompt with a local component that is iteratively decomposed into two low-rank factors and a residual term. The low-rank factors restrict the optimization space and improve generalization and robustness to privacy noise, whereas the residual compensates for information lost during factorization and preserves personalized expressiveness.

Unlike approaches that perform factorization only once, DP-FPL ~\cite{tran2025privacy} repeats the decomposition during every training iteration and reconstructs the local-prompt gradient from the low-rank gradients. Privacy is enforced through a hybrid mechanism: LDP and global differential privacy (GDP). Both mechanisms employ clipped gradients and Gaussian perturbation, with the noise levels selected to satisfy formal ($epslion, \delta$)-LDP and GDP guarantees. The residual is used in the forward computation but is not directly perturbed because it is excluded from the local-prompt gradient reconstruction. 
The results show that the residual term benefits both local and neighboring-class performance, particularly under low-rank and high-noise conditions. They also indicate that moderate privacy noise may act as a regularizer and improve generalization, although excessive noise eventually degrades overall utility. Nevertheless, the evaluation is limited primarily to CLIP-based visual classification, a small set of benchmark datasets, and simulated pathological or Dirichlet client distributions. Moreover, the claimed regularization effect and protective role of the residual term are largely supported empirically rather than through a complete theoretical account.

\subsection{Multimodal Fusion through LLMs and FL}

Pilot \cite{xiong2025pilot} introduces the federated multimodal instruction tuning (FedMIT) task and provides a principled framework for collaboratively fine-tuning multimodal LLMs under severe task and data heterogeneity. Unlike prior federated instruction tuning approaches that focus on unimodal NLP tasks, FedMIT addresses the more challenging setting where clients perform different multimodal tasks (e.g., VQA~\cite{lu2023multi}, captioning, and visual grounding) on decentralized data. To tackle this heterogeneity, Pilot adopts a two-stage ``adapter-on-adapter" design within the vision–language connector: i) a task-specific and client-specific feature mining stage that disentangles task-relevant and personalized visual representations via orthogonality constraints, and ii) a cross-task interaction stage that introduces a cross-task mixture-of-adapters (CT-MoA) module to selectively integrate knowledge from heterogeneous tasks. On the server side, Pilot further proposes task-aware aggregation for visual adapters and an adaptive distance-based aggregation strategy for text adapters, mitigating negative interference across clients.

In federated prompt tuning under non-IID, locally learned prompt sets may encode concepts in arbitrary orders, causing position-wise averaging to combine semantically unrelated prompts into an uninformative global representation. Therefore, Weng~\emph{et al.}~\cite{weng2024probabilistic} propose probabilistic federated prompt tuning (PFPT), which reframes federated aggregation as a distributed set-modeling problem rather than conventional parameter averaging.
PFPT models each client's prompts as samples from a hierarchical generative model parameterized by a server-maintained set of global summarizing prompts. Each round, a client selects a relevant subset of global prompts via a Bernoulli point-process model and fine-tunes them locally with a frozen pretrained vision transformer and a personalized classification head, with resulting local prompts modeled as Gaussian perturbations around their associated summarizing prompts. The server then infers latent local–global prompt associations and aggregates prompts encoding similar contextual information, aligning at the concept level rather than by prompt position.

The resulting mixed discrete–continuous problem is solved via alternating optimization: (1) Generative-model and summarizing-prompt parameters are updated by gradient descent given associations; (2) Association inference reduces to weighted bipartite matching solved via the Hungarian algorithm (cubic complexity in the maximum number of local prompts matched) given parameters. The global prompt pool is non-parametric, unused prompts are pruned. While heterogeneous clients can expand it to represent new concepts.
Nevertheless, evaluation is limited to vision classification with a pretrained ViT under largely simulated heterogeneity and imbalance. In addition, matching cost and pool size may grow with client diversity. The prompt capacity needed for larger pretraining and downstream domain shifts is neither quantified nor theoretically bounded. Meanwhile, despite reduced communication relative to full-model fine-tuning, prompt tuning still requires storing intermediate gradients through the frozen backbone, imposing non-trivial client-side memory overhead.


Rather than proposing a new aggregation algorithm for federated vision-language model adaptation, Wang~\emph{et al.}~\cite{Wang2025empirical} provide a systematic empirical analysis of federated prompt learning under label skew, domain shift, and their coexistence, examining sensitivity to communication rounds, aggregation strategies, client scale, and prompt length.
Using a frozen CLIP model with a ViT-B/16 encoder, the study federatively trains textual or visual prompts via sample-weighted aggregation, evaluated on CIFAR-100 (Dirichlet-based label skew) and Office-Home, Office31, and DomainNet (domain shift). The authors also analyze similarity of optimization directions across clients and between local and global prompts to explain the distinct behaviors of LPT and VPT.
Results show FPL is relatively stable across communication rounds, aggregation rules, client counts, and prompt lengths. Equal aggregation performs slightly better under label skew, while sample-weighted aggregation is preferable under domain shift; more clients modestly reduce performance under label skew but show no consistent effect under domain shift, and longer prompts yield only incremental gains at added computational cost.

However, evidence is confined to CLIP-based image classification with a single ViT-B/16 backbone, four datasets, simulated heterogeneity, fixed configurations, and one random seed; evaluation mainly compares internal prompt variants rather than benchmarking against recent federated prompt-learning methods, and the optimization-direction-similarity explanation remains empirical, without establishing a causal or theoretical link to generalization.

Considering the trade-off between predictive performance and out-of-distribution (OOD) robustness in federated prompt learning for vision-language models, Liao~\emph{et al.}~\cite{Liao2025FOCoOp} design FOCoOp, a federated OOD-aware context optimization framework jointly modeling global in-distribution prompts, client-specific local prompts, and OOD prompts. 

At each client, Bi-level OOD Separations (BOS) optimizes the three prompt sets to establish class-level separation between categories and distribution-level separation between ID and OOD samples. It employs bi-level distributionally robust optimization to perturb global and OOD prompts within optimal-transport uncertainty sets, thereby exposing the model to challenging prompt distributions without requiring real OOD training images. At the server, Global-view OOD Consistency (GOC) aligns aggregated global prompts with client-generated OOD prompts through semi-unbalanced optimal transport. OOD prompts that closely resemble global ID prompts are used to calibrate the latter, whereas distant prompts are retained and redistributed as globally consistent OOD references. 

FOCoOp consistently achieves the strongest overall balance between classification and OOD robustness. Under the ten-client pathological setting, it reaches 93.85\% ID accuracy, 91.47\% covariate-shift accuracy, 19.50\% FPR95, and 95.42\% AUROC on CIFAR-100, and 88.35\% ID accuracy, 83.56\% covariate-shift accuracy, 21.73\% FPR95, and 96.56\% AUROC on TinyImageNet. Ablations confirm BOS and GOC are complementary, with excluding GOC generally causing larger degradation, underscoring the importance of globally consistent ID–OOD discrimination. Similarity visualizations further show strong within-class image prompt alignment versus weaker ID–OOD prompt matching.

Lu~\emph{et al.}~\cite{lu2025fedpob} study black-box prompt optimization for large language models under three practical constraints: expensive model queries, limited access to proprietary model parameters, and the need for privacy-preserving collaboration among multiple users. The authors propose federated prompt optimization via bandits (FedPOB) and a preference-based extension, FedPOB-Pref, which allow multiple agents to search for effective discrete prompts without sharing their local prompt-evaluation histories.
FedPOB encodes discrete prompts using a pretrained text encoder and models prompt performance with a linear reward function. Each agent selects prompts using a LinUCB exploration-exploitation policy and periodically sends summary statistics to a central server, which aggregates and redistributes this information. Communication is event-triggered, meaning synchronization only happens once newly collected information passes a preset threshold. FedPOB-Pref extends this idea to pairwise feedback using federated linear dueling bandits: it picks one prompt for exploitation and another for exploration, learns from Bradley-Terry-Luce preference comparisons, and uses dynamic regularization to reduce optimization drift across heterogeneous agents.

That said, the framework assumes all agents are optimizing prompts for the same task, with heterogeneity captured only through different candidate prompt spaces. Hence, it is unclear how well it would handle agents with different tasks or conflicting preferences. The methods also assume a linear relationship between pretrained prompt embeddings and rewards, and FedPOB-Pref further assumes a Bradley-Terry-Luce preference model; both assumptions may not hold for more complex, non-linear prompt-performance relationships. In addition, preference feedback is simulated from validation scores rather than gathered from real users, so it remains unclear how robust the method is to noisy or manipulated preferences. Finally, while raw prompts and evaluation histories are kept private, agents still send learned statistics to a trusted central server, and the paper does not offer DP, secure aggregation, or any analysis of potential information leakage.

\textbf{Answers to RQ2:} FPL approaches exhibit systematic trade-offs across performance, communication efficiency, computational overhead, scalability, personalization, and heterogeneity handling, with no single method dominating on all dimensions simultaneously. 
The survey's unified benchmark (Section \ref{sec:Benchmarking})shows that LoRA achieves the highest and most stable accuracy across IID and non-IID settings but incurs substantially larger communication costs, whereas prompt tuning minimizes communication and trainable parameters at the expense of lower and more variable accuracy. Notably, P-tuning's larger parameter count does not translate into greater stability, indicating that parameterization structure matters more than raw scale for robustness under heterogeneous optimization. Methods that improve personalization through richer mechanisms (e.g., multi-expert prompt mixtures or global-local prompt decomposition) gain adaptability at the cost of increased communication, storage, and computation. While aggressive local adaptation risks eroding the generalization inherited from pretrained backbones, a tension partially mitigated through low-rank restrictions and contrastive regularization. Similarly, naive position-wise prompt averaging often fails under non-IID data due to semantic misalignment across clients, motivating concept-level or attention-based aggregation strategies that improve heterogeneity handling but introduce additional computational overhead and remain validated mainly under simulated conditions.

\section{Security and Privacy Issues in FPL}\label{sec:SecurityPI}

\subsection{Security Issues in FPL}

Safety alignment can be severely compromised by a simple, low-cost, and highly stealthy data poisoning attack in federated instruction tuning (FedIT) of LLMs~\cite{ye2024emerging}. Unlike traditional FL poisoning attacks, the proposed attack does not introduce conflicting optimization signals. Instead, malicious clients are trained on harmful instruction: response pairs. These pairs remain optimization-consistent with benign training data. As a result, the attack exploits the intrinsic alignment objective of instruction tuning. This makes existing model-level FL defenses largely ineffective. Extensive experiments demonstrate that such attacks can reduce safety metrics by up to 70\%, while classical robust aggregation methods provide only marginal protection. To address this gap, the authors propose a post-hoc, server-side defense strategy that decouples safety recovery from client-side training by automatically generating aligned and normal instruction–response data and performing lightweight fine-tuning on the aggregated model. This defense is shown to be plug-and-play, scalable, and capable of restoring safety performance by up to 69\% without significantly degrading helpfulness. 

Zheng et al.~\cite{zheng2024safely} propose FL-GLM, a secure and efficient split-learning-based federated learning framework for LLMs, addressing the incompatibility of conventional FL (e.g., FedAvg) with the heavy computation and privacy risks of large models. The framework places the embedding layer and the first and last transformer blocks on the client, keeping intermediate blocks on the server to prevent embedding-gradient leakage, while encrypting all client–server ``smashed data" via public-key encryption to guard against peer-client eavesdropping. To mitigate split learning's low efficiency, FL-GLM introduces client-batch and server-hierarchical parallelism, cutting training time by over 48\% versus serial execution. While limited to ChatGLM-scale models and still incurring communication overhead, FL-GLM offers a practical blueprint showing that encrypted, parallelized split architectures can jointly reconcile utility, efficiency, and privacy in federated LLM training.

Huang et al.~\cite{huang2024fast} propose a fast, accurate, and security-preserving distributed training framework for LLMs that addresses parameter and data leakage risks in federated settings. Motivated by the limitations of prior defenses: DP's accuracy loss and the heavy overhead of multi-party computation, the authors design a model slicing-based secure architecture integrating trusted execution environments (TEEs) with lightweight encryption, via two complementary schemes (i) \textit{Method 1}; (ii) \textit{Method 2}.
\textit{Method 1} targets consumer-grade, small-memory TEEs (e.g., Intel SGX) by isolating sensitive fine-tuned components (LoRA or P-Tuning v2 embeddings) within client- and server-side TEEs, protecting intermediate activations exchanged with GPUs via one-time-pad encryption.\textit{Method 2} improves efficiency and accuracy further by splitting the LLM by layers, deploying deeper layers in a large-memory server-side TEE (Intel TDX/SGX) while freezing client-side layers.  A novel sparsification parameter fine-tuning (SPF) strategy selectively updates key attention heads and combines LoRA for MLP layers.

FedShield-LLM~\cite{mia2025fedshield} is a secure, scalable federated fine-tuning framework for LLMs that addresses the trade-off between strong privacy protection and high model utility in cross-silo FL. Motivated by the vulnerability of standard federated LLM fine-tuning to inference attacks (e.g., gradient inversion, membership inference) and the performance degradation caused by noise-based defenses like DP, FedShield-LLM integrates PEFT (LoRA) with unstructured pruning and Fully Homomorphic Encryption (FHE). Clients fine-tune only low-rank LoRA adapters on local data, prune less informative adapter weights to reduce attack surface and communication cost, and encrypt updates via the CKKS scheme, enabling the server to aggregate encrypted parameters without accessing plaintext.
The authors provide theoretical convergence guarantees, showing the encrypted, sparsified FedAvg procedure retains sublinear convergence, and formal security arguments demonstrating resistance to gradient inversion and inference attacks under standard cryptographic assumptions. Experiments on LLaMA-2 (7B and 13B) across medical, financial, mathematical, and general instruction-tuning benchmarks demonstrate FedShield-LLM consistently outperforms Vanilla FL and DP-LoRA in convergence speed, training stability, and text quality, achieving BERTScore gains while approaching the response quality of GPT-4-class models.

However, in federated graph learning, the participants may differ not only in graph structures and feature distributions but also in downstream tasks, including node-, edge-, and graph-level prediction. To address this multifaceted heterogeneity in federated graph learning, Guo\emph{et al.}~\cite{guo2024against} propose FedGPL, an asymmetric federated graph prompt-learning framework that separates universal graph representation learning from client-specific prompting and prediction. FedGPL integrates a server-hosted GNN with locally maintained graph prompts and task heads. 
Meanwhile, the hierarchical directed transfer aggregator (HiDTA) estimates directional transferability between clients and performs personalized aggregation to encourage beneficial cross-task knowledge transfer while suppressing potentially harmful contributions. The Virtual Prompt Graph (VPG) adaptively adds informative virtual structures and removes or neutralizes redundant nodes and edges, thereby preserving domain-specific patterns while reducing graph-data divergence. FedGPL additionally employs split learning and differential-privacy noise to protect the graph embeddings and gradients exchanged between clients and the server. The principal contribution therefore lies in jointly addressing graph-task and graph-data heterogeneity through asymmetric transferability-aware aggregation and adaptive structural prompting, while retaining parameter and communication efficiency. The accompanying theoretical analysis provides conditional support for heterogeneity reduction: HiDTA requires positive bidirectional transferability, whereas the VPG analysis assumes particular graph-representation distributions and similar prompting ratios across clients. Nevertheless, the empirical evaluation is restricted to five benchmark graph datasets, generally uses three clients per task level, assumes full participation and one local training step per round, and adopts a semi-honest threat model. Moreover, stronger privacy constraints reduce predictive accuracy, indicating a non-trivial privacy–utility trade-off.

Khan~\emph{et al.}~\cite{khan2026sabre} address the largely unexplored security risks of FPL, showing that communication-efficient prompt updates remain vulnerable to targeted backdoor attacks despite a frozen vision-language backbone. It introduces a learnable, visually imperceptible noise trigger that malicious clients inject into local images to shift their CLIP representations toward an attacker-chosen target class, preserving accuracy on clean inputs while causing systematic misclassification of triggered samples. To counter this, the paper proposes selective and accurate backdoor rejection for federated prompt learning (SABRE-FL), a lightweight server-side defense that detects and removes poisoned client updates via embedding-space anomaly detection.

The attack adapts the centralized BadCLIP mechanism to a federated, prompt-only setting: each compromised client jointly optimizes a local prompt and an additive noise trigger, relabeling triggered images to a fixed target class. Although the CLIP encoders stay frozen, the trigger shifts image embeddings toward the target-class text representation, and the resulting malicious prompt updates propagate through FedAvg aggregation. Under the default threat model, the adversary controls 25\% of clients, knows the public backbone and training process, and can modify local data and labels but not server aggregation directly.

The theoretical argument also depends on triggered and clean embeddings maintaining a stable separation margin and on the auxiliary detector achieving near-zero training error, so robustness against adaptive attackers minimizing this separation, alternative trigger types, and direct model-poisoning attacks remains unestablished. The method further assumes a boundable number of malicious clients and is evaluated mainly with a single CLIP ViT-B/16 backbone, with the authors flagging model poisoning, broader threat models, and additional vision-language architectures as directions for future work.

\subsection{Privacy Issues in FPL}

Luo \emph{et al.} \cite{luo2024fine}
systematically examine FL as a privacy-preserving paradigm for fine-tuning LLMs in automated program repair (APR), motivated by the growing reliance of LLM-based repair methods on high-quality industrial code that cannot be centrally shared due to confidentiality constraints. Using a private industrial dataset (TutorCode) for training and the leakage-resistant EvalRepair-Java benchmark for evaluation, the authors investigate three core challenges: whether federated fine-tuning can effectively enhance repair performance, how code heterogeneity (feature-skewed Non-IID data) influences generative repair tasks, and how different FL algorithms impact optimization. By adopting parameter-efficient fine-tuning (QLoRA), the framework significantly reduces communication and memory overhead while preserving data locality. Extensive experiments across six state-of-the-art code LLMs show that federated fine-tuning consistently and substantially improves bug-fixing performance over local fine-tuning and often approaches, or even surpasses, centralized training, with gains of up to 16.6\% in Top@10 and 18.4\% in Pass@10. Surprisingly, heterogeneous code distributions do not degrade performance and can even enhance generalization, highlighting the robustness of LLMs in feature-skewed federated generative tasks. While FedAvg emerges as the most stable and effective aggregation strategy, personalized FL remains challenging for LLM-based repair.

ROFED-LLM~\cite{wang2025rofed} is a robust, privacy-preserving federated learning framework for training LLMs in adversarial wireless environments, addressing the vulnerability of conventional federated LLM training to jamming attacks, device heterogeneity, and privacy leakage. It integrates split federated learning with a multi-modal defense spanning both model and communication layers: on the learning side, LLMs are partitioned between clients and server, with local differential privacy (LDP), secure aggregation, and dynamic parameter pruning reducing information leakage and computational burden; on the communication side, adaptive beamforming, jamming detection, and resource allocation optimization enable real-time mitigation of interference and adversarial attacks over wireless channels.
The authors provide convergence guarantees and privacy analysis, showing ROFED-LLM achieves sublinear convergence while maintaining formal DP. Experiments across diverse natural language processing (NLP) benchmarks (e.g., general language understanding evaluation benchmark (GLUE)~\cite{wang2018glue}, SQuAD, WikiText-103, FinancialPhraseBank) show it consistently outperforms FedAvg, FedProx, SCAFFOLD, and SplitFed. ROFED-LLM prioritizes FedLLM methods with a 47\% lower privacy leakage and significantly greater robustness under severe jamming, with only marginal communication overhead.

\section{Experimental Benchmarking and Empirical Insights}\label{sec:Benchmarking}

Besides of the systematic analysis for the existing work on the FPL topic, we conduct a unified empirical benchmark to complement the methodological taxonomy with reproducible quantitative evidence. A major difficulty in this area is that published results are often obtained under disparate task formulations, model backbones, data partitions, and evaluation protocols, rendering direct comparison unreliable. Therefore, rather than pursuing state-of-the-art performance, this benchmark evaluates representative federated PEFT strategies under a strictly consistent experimental protocol, thereby enabling a fair and controlled comparison across methods originally tested in heterogeneous settings.

\subsection{Unified Benchmark Evaluation}
\label{sec:core_benchmark_results}

Table~\ref{tab:core_benchmark_results} reports the core benchmark results on AG News using TinyLlama-1.1B-Chat as the backbone. All methods were trained with four clients over 100 communication rounds and evaluated on the same held-out test set using the post hoc candidate-label scoring protocol. We report the mean and standard deviation over three random seeds.

\begin{table*}
\centering
\footnotesize
\caption{
Unified benchmark results of federated PEFT methods on AG News using
TinyLlama-1.1B-Chat. Values are reported as mean $\pm$ standard deviation
over three random seeds.
}
\label{tab:core_benchmark_results}
\setlength{\tabcolsep}{4.2pt}
\renewcommand{\arraystretch}{1.10}

\begin{threeparttable}
\begin{tabular}{
    ll
    cc
    cc
    cc
}
\toprule

\multirow{2}{*}{Setting}
&
\multirow{2}{*}{Method}
&
\multicolumn{2}{c}{Predictive Performance}
&
\multicolumn{2}{c}{Adaptation Scale}
&
\multicolumn{2}{c}{System Efficiency}
\\

\cmidrule(lr){3-4}
\cmidrule(lr){5-6}
\cmidrule(lr){7-8}

&
&
Acc. (\%) $\uparrow$
&
Test Loss $\downarrow$
&
Params (M) $\downarrow$
&
Trainable (\%) $\downarrow$
&
Download (MB) $\downarrow$
&
FL Time (min) $\downarrow$
\\

\midrule

\multirow{3}{*}{IID}
& LoRA
& $88.78 \pm 0.63$
& $0.091 \pm 0.001$
& 1.13
& 0.1023
& 5.04
& $76.80 \pm 0.53$
\\

& Prompt tuning
& $82.57 \pm 7.28$
& $0.133 \pm 0.048$
& 0.20
& 0.0186
& 0.11
& $110.23 \pm 0.84$
\\

& P-tuning
& $81.74 \pm 1.89$
& $0.115 \pm 0.009$
& 12.63
& 1.1351
& 0.39
& $82.46 \pm 0.89$
\\

\addlinespace[2pt]
\midrule

\multirow{3}{*}{LDA $\alpha=0.1$}
& LoRA
& $88.85 \pm 0.12$
& $0.093 \pm 0.002$
& 1.13
& 0.1023
& 5.04
& $78.55 \pm 3.19$
\\

& Prompt tuning
& $83.43 \pm 3.85$
& $0.142 \pm 0.049$
& 0.20
& 0.0186
& 0.11
& $111.80 \pm 5.85$
\\

& P-tuning
& $77.56 \pm 13.24$
& $0.151 \pm 0.049$
& 12.63
& 1.1351
& 0.39
& $89.19 \pm 13.00$
\\

\addlinespace[2pt]
\midrule

\multirow{3}{*}{LDA $\alpha=0.5$}
& LoRA
& $88.95 \pm 0.35$
& $0.091 \pm 0.002$
& 1.13
& 0.1023
& 5.04
& $86.95 \pm 13.08$
\\

& Prompt tuning
& $83.10 \pm 6.28$
& $0.145 \pm 0.077$
& 0.20
& 0.0186
& 0.11
& $113.90 \pm 6.48$
\\

& P-tuning
& $73.45 \pm 18.28$
& $0.116 \pm 0.014$
& 12.63
& 1.1351
& 0.39
& $82.23 \pm 0.44$
\\

\addlinespace[2pt]
\midrule

\multirow{3}{*}{LDA $\alpha=1.0$}
& LoRA
& $88.76 \pm 0.08$
& $0.094 \pm 0.004$
& 1.13
& 0.1023
& 5.04
& $78.94 \pm 3.62$
\\

& Prompt tuning
& $79.69 \pm 11.77$
& $0.142 \pm 0.054$
& 0.20
& 0.0186
& 0.11
& $113.43 \pm 3.91$
\\

& P-tuning
& $83.65 \pm 3.63$
& $0.116 \pm 0.009$
& 12.63
& 1.1351
& 0.39
& $89.56 \pm 11.30$
\\

\bottomrule
\end{tabular}

\begin{tablenotes}
\footnotesize
\item Test Loss is the server-side weighted-average test loss reported by
the federated framework.
Download denotes the average download volume per worker reported by the
framework-level system monitor; the original values are converted to MB
for consistent presentation.
All experiments use four clients, full client participation, 100
communication rounds, and the same held-out evaluation set.
\end{tablenotes}

\end{threeparttable}
\end{table*}

Several consistent observations emerge from the benchmark results. First, LoRA provides the most stable performance among the compared methods. Its accuracy remains around 88--89\% across IID and all LDA-based non-IID settings, with consistently small standard deviations. These results suggest that LoRA is comparatively robust to the evaluated data partition changes, although it incurs a larger communication payload than prompt-based methods.

Second, prompt tuning achieves the lowest trainable parameter ratio and the smallest average download volume, but its predictive performance is more sensitive to random seed and data partition. This suggests that prompt tuning provides a strong communication-efficiency advantage, but its optimization stability is weaker than LoRA under the current federated setting.

Third, P-tuning does not consistently outperform prompt tuning despite using substantially more trainable parameters. In particular, it exhibits large variance under LDA $\alpha=0.1$ and $\alpha=0.5$, indicating that increasing the prompt-encoder parameter scale does not automatically translate into more stable federated adaptation. This observation is important because it suggests that the practical value of a parameter-efficient method should be judged jointly by accuracy, stability, trainable scale, and communication behavior rather than by parameter count alone.

\subsection{Empirical Analysis of Performance, Robustness, and Reproducibility}

\label{sec:preliminary_observations}

\begin{figure*}[h]
\centering

\subfloat[]{
    \includegraphics[width=0.315\textwidth]{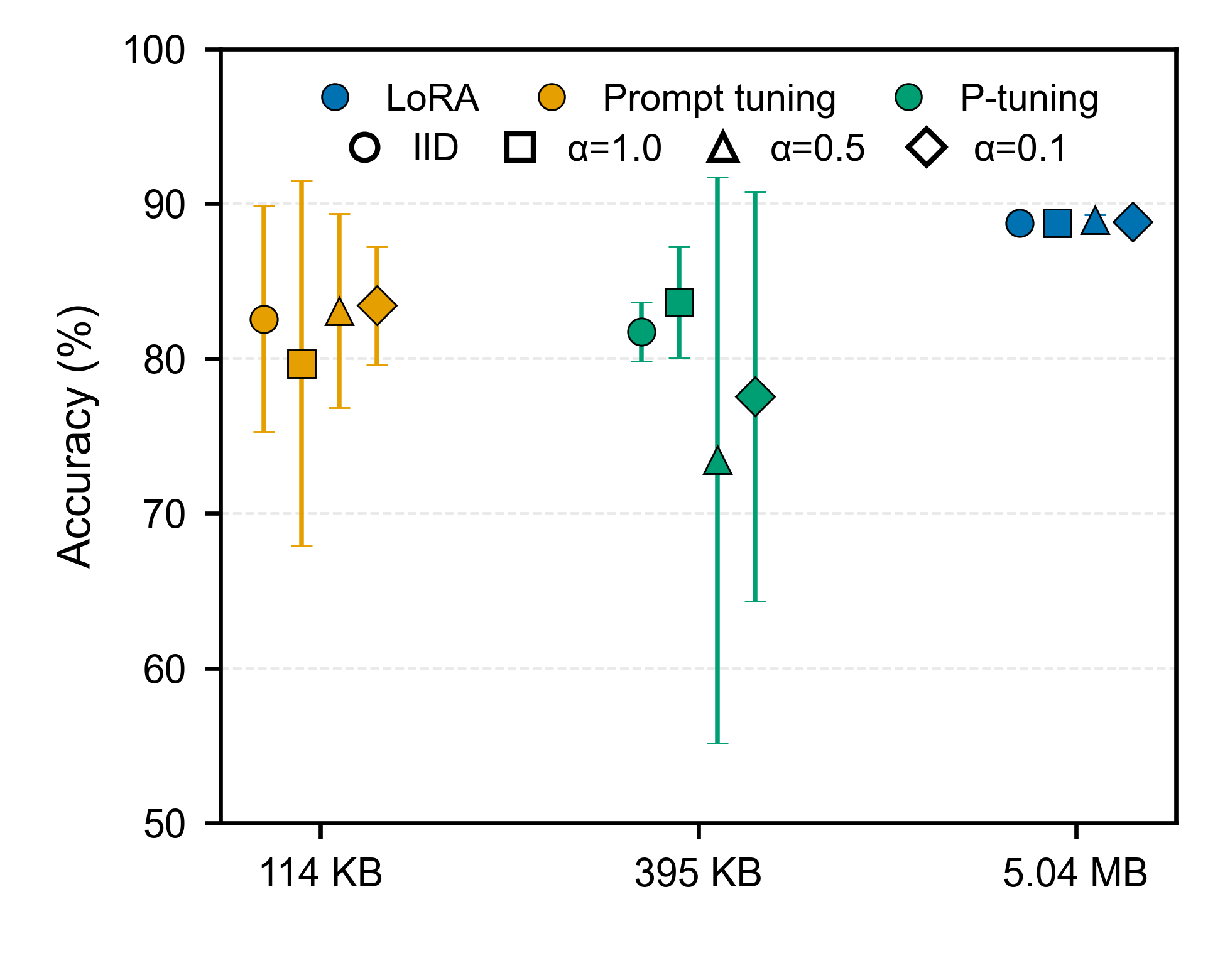}
    \label{fig:tradeoff}
}
\hfill
\subfloat[]{
    \includegraphics[width=0.315\textwidth]{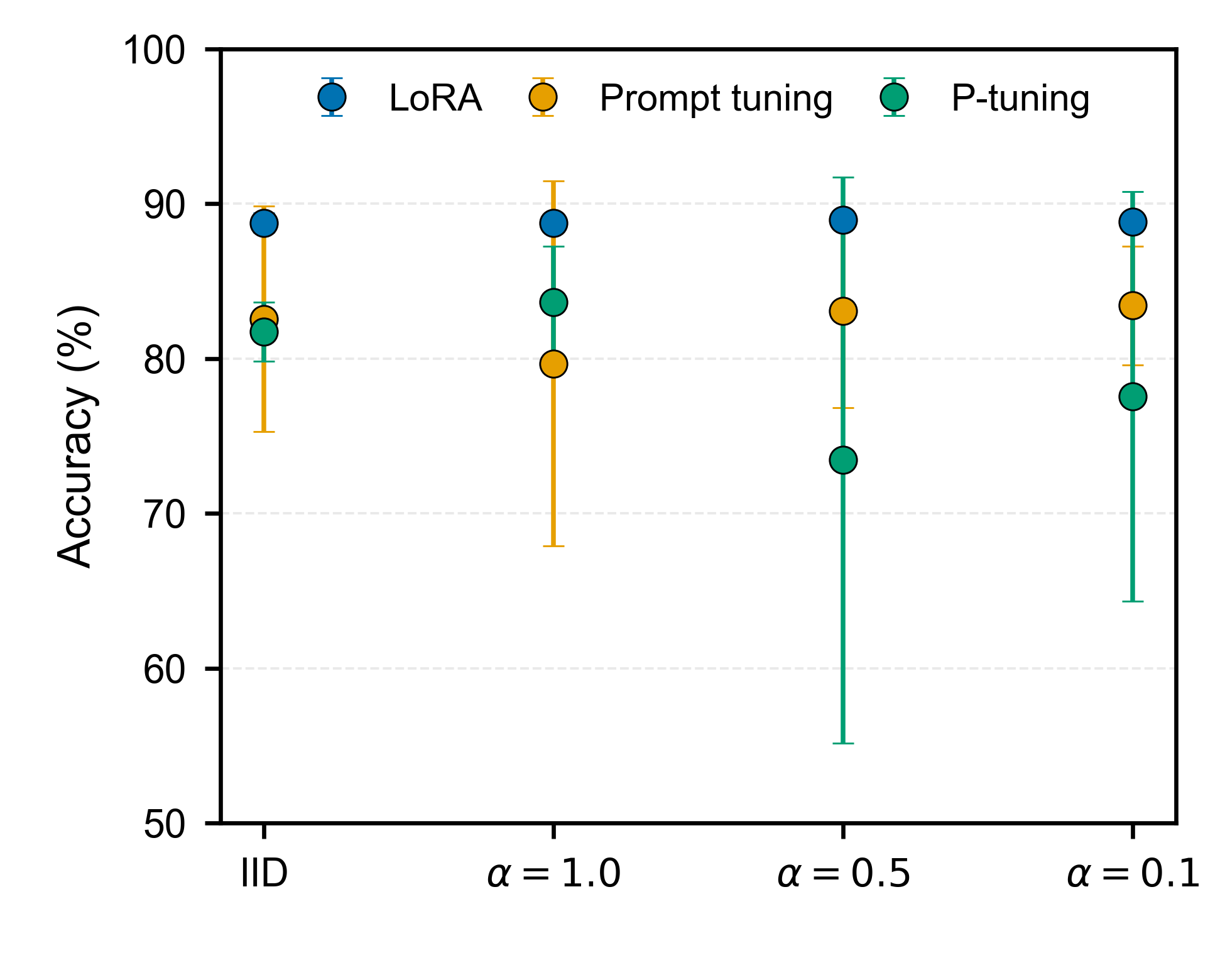}
    \label{fig:heterogeneity}
}
\hfill
\subfloat[]{
    \includegraphics[width=0.315\textwidth]{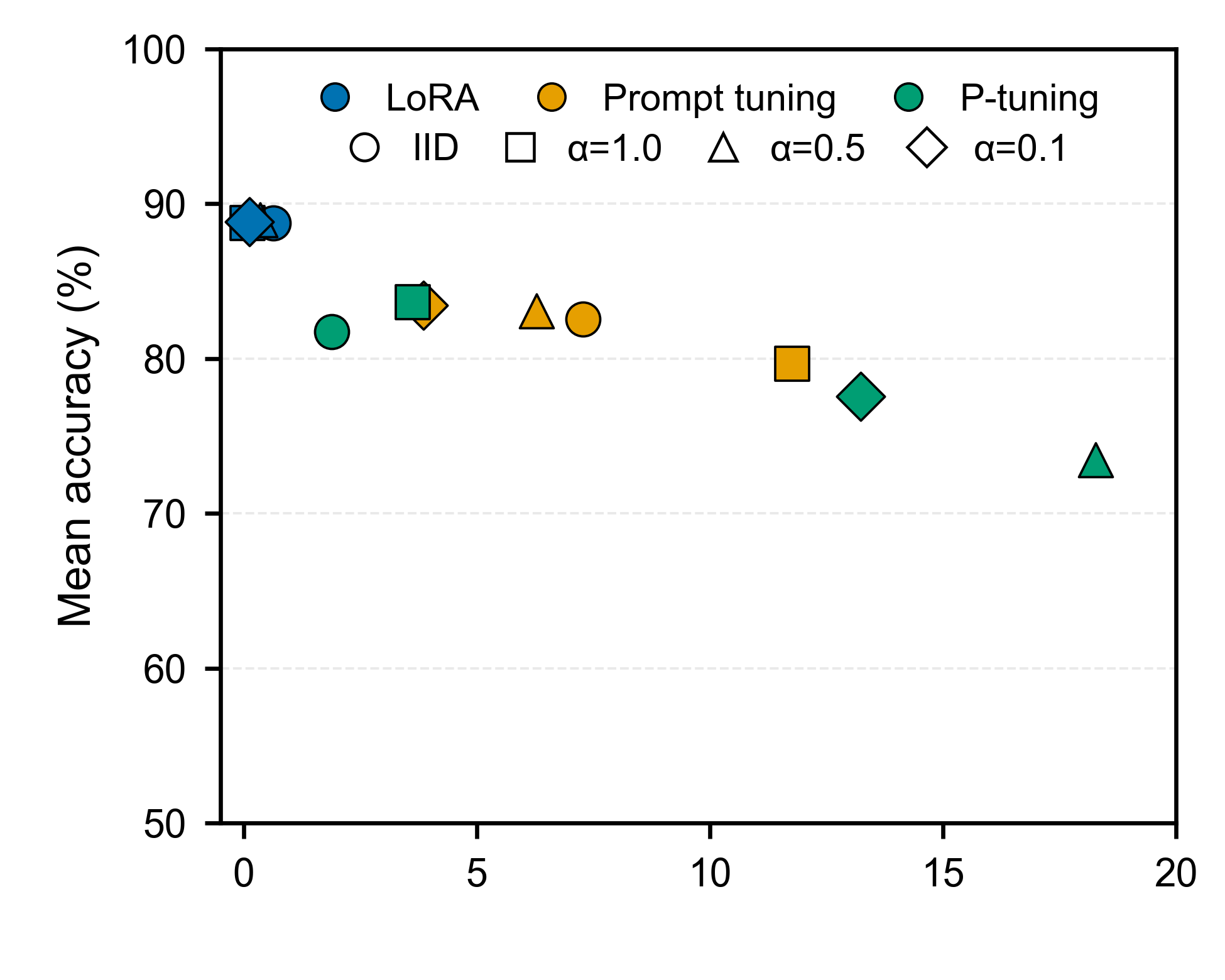}
    \label{fig:stability}
}

\caption{
Comparison of representative federated PEFT methods under the unified benchmark from three complementary perspectives.
(a) Performance--efficiency trade-off between predictive accuracy and communication cost.
(b) Robustness under IID and progressively heterogeneous federated data partitions.
(c) Reproducibility across random seeds, where each point denotes one method under one partition setting and is positioned according to its mean accuracy and cross-seed standard deviation.
Error bars indicate the standard deviation over three random seeds.
}
\Description{Three benchmark plots compare LoRA, prompt tuning, and P-tuning. Panel (a) plots accuracy against communication volume; LoRA has the highest accuracy at the largest communication volume, prompt tuning has the smallest volume, and P-tuning is intermediate with larger variance. Panel (b) shows accuracy across IID and progressively heterogeneous LDA partitions, where LoRA remains stable while P-tuning degrades and varies more under stronger heterogeneity. Panel (c) plots mean accuracy against cross-seed standard deviation, with LoRA concentrated at high accuracy and low variance.}
\label{fig:benchmark_overview}

\end{figure*}

Fig.~\ref{fig:benchmark_overview} complements the quantitative results in Table~\ref{tab:core_benchmark_results} by visualizing the empirical characteristics of representative federated PEFT methods from three complementary perspectives, namely communication efficiency, robustness under heterogeneous federated settings, and reproducibility across random seeds. Several important empirical observations can be derived from the benchmark results. (1) The compared methods exhibit substantially different robustness profiles under federated heterogeneity. Among all evaluated methods, LoRA consistently demonstrates the strongest stability across IID and non-IID settings. Its task accuracy remains highly stable across different random seeds and LDA partition conditions, while its test loss also remains comparatively well-controlled. This behavior highlights that low-rank parameter adaptation provides a relatively robust optimization path for federated LLM fine-tuning, even when client distributions become heterogeneous.

(2) Communication-efficient methods do not necessarily preserve stable predictive behavior under federated optimization. Prompt tuning achieves the smallest trainable parameter ratio and the lowest communication volume among all evaluated methods. However, its performance variance becomes noticeably larger under certain non-IID settings, indicating that extreme parameter compression may increase optimization sensitivity in federated environments. In particular, several prompt-tuning runs show substantial seed-dependent fluctuations despite using the same backbone model and evaluation protocol.

(3) Larger trainable parameter scales do not automatically guarantee stronger robustness. Although P-tuning introduces considerably more trainable parameters than prompt tuning, it exhibits the greatest instability across several heterogeneous settings, including occasional performance collapse with moderate non-IID partitions. This observation suggests that federated prompt learning performance is influenced not only by the scale of trainable parameters but also by how the parameterization interacts with heterogeneous local optimization dynamics.

In summary, our benchmark demonstrates that federated PEFT methods must be evaluated as a multi-objective design problem rather than through predictive accuracy alone. Specifically, these findings suggest that communication-efficient adaptation should not be evaluated solely by parameter reduction; indeed, a strategy with superior parameter efficiency may suffer from heightened optimization instability, whereas a more stable method may incur substantially larger communication overhead. Consequently, practical federated deployment requires balancing these distinct trade-offs—spanning effectiveness, optimization robustness, parameter footprint, and communication cost—all of which constitute complementary dimensions in the design space of federated prompt learning instead of being optimized in isolation.

\section{Insights and Future Research}

Although we have reviewed many related studies, the integration of FL and LLMs remains at an early stage of maturity when judged against the recurring limitations acknowledged throughout the surveyed studies: simulated heterogeneity, small or moderate client populations, single-backbone evaluation, informally justified privacy claims, and fragmented experimental protocols. These limitations become sharper still as the field moves from static model adaptation toward federated agentic systems, where models act, retrieve, and communicate autonomously rather than simply producing text. Synthesizing these gaps, we organize future research into nine directions spanning evaluation, theory, systems, and increasingly central trustworthiness.

\subsection{Standardized, Reproducible, and Multi-Objective Evaluation}

As our unified benchmark in Section~\ref{sec:Benchmarking} demonstrates, methods with comparable accuracy can differ sharply in stability, communication cost, and seed-to-seed variance—dimensions that most published results, obtained under disparate backbones, partitions, and protocols, do not jointly report. Future work should move toward shared benchmark suites with multi-seed reporting and standardized non-IID partition generators, so that communication efficiency, robustness, and parameter footprint are treated as co-equal objectives rather than being subordinated to a single accuracy figure.

\subsection{Theoretical Foundations for Federated Prompt and PEFT Optimization}
Convergence guarantees remain largely confined to simplified regimes (e.g., PromptFL's $\mathcal{O}(1/\sqrt{T})$ bound \cite{guo2023promptfl}), while empirical phenomena such as the generalization–personalization trade-off in FedPGP~\cite{3692070.3692451}, the concept-drift dynamics in PFPT~\cite{weng2024probabilistic}, or the ID–OOD separation margin in FOCoOp~\cite{Liao2025FOCoOp} are supported only empirically. A principled theory connecting prompt rank, capacity, and domain shift to generalization bounds under non-IID, nonlinear transformer dynamics is still missing, and would let practitioners choose prompt configurations analytically rather than by grid search.

\subsection{Heterogeneity- and Resource-Aware Optimization at Realistic Scale}

FlexLoRA \cite{bai2024federated}, FAH-QLoRA \cite{gao2025federated}, and Fed-HeLLo~\cite{zhang2025fed} all address resource heterogeneity, but under simulated conditions, moderate model sizes, and limited client counts. Real deployments will involve thousands of genuinely heterogeneous devices and clients pursuing different tasks—not merely different label or domain distributions—so validating these mechanisms on larger LLM backbones and authentic edge testbeds is a necessary next step.

\subsection{ LLMs as Intelligent Controllers of Federated Systems}
Section IV's LLM-enhanced FL largely covers data augmentation and pseudo-labeling; a less explored direction is using LLMs as adaptive controllers—for client selection, aggregation-weight design, hyperparameter tuning, and anomalous-update detection—closing the loop between ``FL for LLMs" and ``LLMs for FL" within a single self-improving system.

\subsection{Federated Agentic Ecosystems: Retrieval, Multi-Agent Collaboration, and Verifiable Trust}
Federated RAG and federated multi-agent systems were introduced in Section I as part of the synergistic-integration category, yet neither is represented by a dedicated study in Section V-B, marking a clear literature gap. Open problems include collaboratively optimizing retrievers or knowledge representations across clients that each retain private knowledge bases, and designing protocols for LLM agents to share task experience or tool-use knowledge without exposing local memory or data. As such systems mature, trust cannot rest on single-round update verification alone: unlike the anomaly-detection defenses developed for prompt-only settings such as SABRE-FL~\cite{khan2026sabre} and FedGPL~\cite{guo2024against}, an agent's trustworthiness must be assessed over extended, multi-step interaction trajectories, where one manipulated tool call or retrieved document can silently corrupt downstream reasoning. Future work should therefore pair federated RAG and multi-agent architectures with trajectory-level verification mechanisms—potentially combining lightweight cryptographic attestation, verifiable computation, or provenance logging—that certify agent behavior without requiring centralized visibility into private client data.

\subsection{Privacy Protection Beyond Training: From Formal Guarantees to Inference-Time and Agentic Leakage}
Several methods justify privacy only indirectly—e.g., PLAN~\cite{gong2026federated} argues that prompts are a compact, less data-revealing medium without formal leakage bounds, and SABRE-FL~\cite{khan2026sabre} requires exchanging raw embeddings without quantifying the resulting risk. Even DP-FPL~\cite{tran2025privacy}, despite offering formal $(\ epsilon, \delta)$ guarantees, evaluates them against a single shadow-model attack. This gap is sharpened by the theoretical result of Nikolaou et al. \cite{nikolaou2025language} that decoder-only Transformers are almost surely invertible, implying that exchanged hidden representations may be reconstructible in principle. The risk extends beyond training: agentic deployments introduce persistent memory, retrieved documents, and inter-agent messages exchanged at inference time, which can leak sensitive local context even when model weights stay private. Future work should derive leakage bounds specific to prompt-, embedding-, and memory-space exchange, and test defenses—privacy-preserving memory management, encrypted or federated retrieval—against adaptive, reconstruction-aware adversaries rather than fixed, known attacks.

\subsection{Safety Alignment and Accountability in Federated and Agentic Deployments}

Ye et al.~\cite{ye2024emerging} show that safety alignment achieved during centralized instruction tuning can be substantially eroded by a small fraction of poisoned federated clients, with conventional robust aggregation providing only marginal protection. This risk is amplified in agentic settings, where a misaligned or jailbroken local model may issue harmful tool calls or propagate unsafe behavior to other agents, rather than merely producing an unsafe text response. Extending post-hoc safety-recovery strategies—such as the server-side realignment fine-tuning in ~\cite{ye2024emerging}, to agentic action spaces, and moving from one-time post-training checks to continuous, round-wise safety auditing, remain open problems. This challenge is compounded by an accountability gap: because no single party observes the complete federation-wide interaction trace, attributing a faulty or harmful agent decision to a specific client, prompt, or aggregation step is substantially harder than in centralized settings, particularly in regulated domains such as healthcare (FedMRG~\cite{che2025llm}) and IoT management (Otoum~\cite{otoum2025llms}) already surveyed in this paper. Future research should explore lightweight, privacy-compatible audit trails and explanation mechanisms that preserve data locality while still enabling post-hoc accountability and regulatory compliance.

\subsection{Incentive Design and Real-World Deployment}

Most surveyed methods assume full or voluntary client participation, yet clients holding valuable proprietary data or compute have limited reason to contribute without appropriate incentives. Coupling incentive-mechanism design with energy- and communication-aware scheduling, and validating results through real-world case studies rather than simulation alone, is needed to translate the substantial methodological progress reviewed in this survey, together with the trustworthiness safeguards outlined above, into deployable, accountable federated LLM systems.

\textbf{Answers to RQ3:} Despite substantial methodological progress, federated prompt learning continues to face significant security, privacy, robustness, and system challenges. On the security front, federated instruction tuning remains vulnerable to stealthy data-poisoning attacks that erode safety alignment while evading conventional robust aggregation, and prompt-only systems remain susceptible to backdoor attacks via imperceptible triggers even when the backbone stays frozen, with existing defenses (e.g., SABRE-FL, FedGPL) yet to be validated against adaptive attackers or broader threat models. Privacy protections are often justified only informally—relying on the implicit assumption that prompts are less data-revealing—rather than through formal guarantees, and even methods offering DP bounds (e.g., DP-FPL) are tested against limited attack models; this gap is compounded by theoretical evidence that transformer representations may be invertible in principle, and by largely unexamined leakage risks arising from persistent memory and inter-agent communication in agentic deployments. Robustness remains constrained by simulated (rather than real-world) heterogeneity, small client populations, single-backbone evaluation, and fragmented experimental protocols, while accountability is undermined by the absence of any single party observing the complete federation-wide interaction trace. Synthesizing these gaps, the survey outlines future directions spanning standardized multi-objective evaluation, theoretical foundations connecting prompt capacity to generalization under non-IID conditions, heterogeneity- and resource-aware optimization at realistic scale, LLMs as intelligent controllers of federated systems, trustworthy federated agentic ecosystems with trajectory-level verification, privacy protection extending beyond training to inference-time and memory-space leakage, continuous safety auditing and accountability mechanisms for agentic deployments, and incentive design coupled with real-world validation—collectively underscoring that the field remains at an early stage of maturity as it transitions from static model adaptation toward autonomous, agentic federated systems.

\section{Conclusion}\label{sec:conclusions}

This survey systematically reviewed federated prompt learning across model adaptation, personalization, inference, applications, and security. Existing studies show that prompts and lightweight adapters can substantially reduce the communication, computation, and storage costs of federated LLM fine-tuning while keeping raw data local. However, methods differ considerably in their ability to handle non-IID data, resource heterogeneity, personalization, scalability, and privacy.
The unified benchmark further indicates that federated PEFT should be evaluated as a multi-objective problem. LoRA generally offers more stable performance, whereas prompt-based methods achieve lower communication costs but may be more sensitive to heterogeneous optimization.
Future research should focus on resource-aware and adaptive optimization, stronger personalization and generalization, standardized evaluation, and formal protection against privacy leakage, poisoning, backdoor, and reconstruction attacks. 

\begin{acks}
This work was partially supported by National Natural Science Foundation of China under Grant No.U25B2030; National Natural Science Foundation of China under Grant No.T2522011; GuangDong Basic and Applied Basic Research Foundation under Grant No. 2025B1515020022, 2026A1515010183.
\end{acks}

\bibliographystyle{ACM-Reference-Format}
\bibliography{references}

\end{document}